%% file: main.tex
\documentclass[10pt]{article}
\PassOptionsToPackage{dvipsnames,svgnames,x11names}{xcolor}
\usepackage[citestyle=authoryear]{colab}
\hypersetup{pdftitle={When Should the Count Change? Learning State Maintenance for Causal Video Counting}}
\colabrunningtitle{StaMina: State Maintenance for Causal Video Counting}
\labname{Colab}
\colabdate{2026-09-28}
\paperurl{https://buaa-colalab.github.io/StaMina/}
\githuburl{}
\title{When Should the Count Change? Learning\\State Maintenance for Causal Video Counting}
\author{Pengyiang Liu$^{1,*}$ \quad Dongyue Lyu$^{1,*}$ \quad Junbo Niu$^{2,*}$\\
Zhongyue Shi$^{1}$ \quad Jiahao Xie$^{1}$ \quad Si Liu$^{1}$\\[0.2em]
\textsuperscript{1}Beihang University \quad \textsuperscript{2}Peking University\\
\textsuperscript{*}Equal contribution}
\graphicspath{{figures/}}
\input{style/math_commands.tex}

\input{management/generated/numbers.tex}

\newif\ifChinese
\Chinesefalse
\input{shared/layout.tex}

\begin{document}
\maketitle
\vspace{-12pt}
\makeatletter
\begin{colababstract}
\input{sections/00_abstract}

\vspace{0.6em}
{\small\textbf{Project Page:}~\url{\@paperurl}}
\end{colababstract}
\makeatother
\vspace{0.5em}
\begin{center}
\begin{minipage}{\textwidth}
\centering
\includegraphics[width=0.90\linewidth]{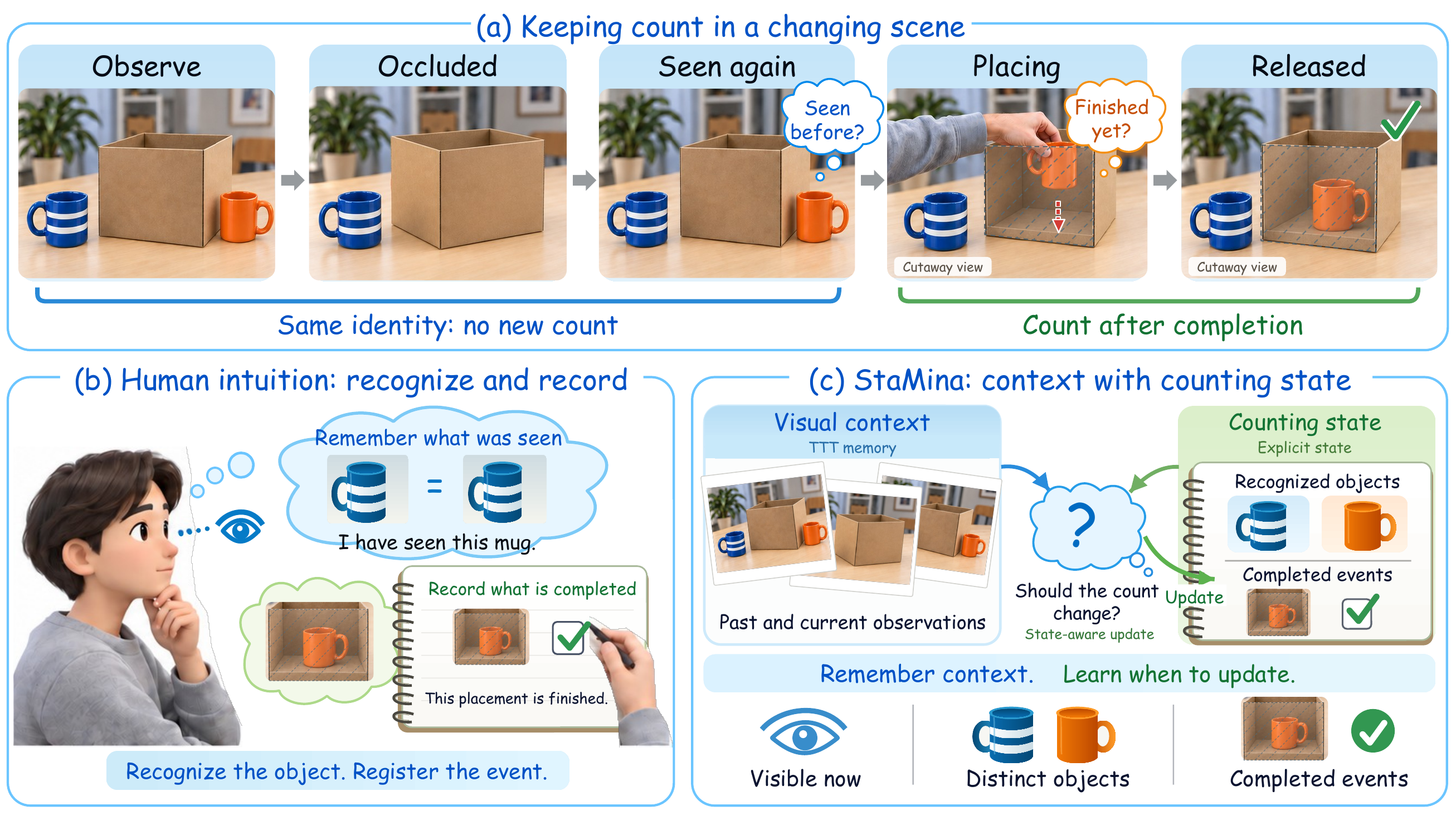}
\captionof{figure}{\textbf{Counting while packing.} (a) Reappearance preserves object identity; lowering and releasing the orange mug distinguish an ongoing placement from a completed event. Hatched cutaways show the placement stages. (b) The inventory clerk uses remembered appearance to recognize objects and a notebook to track completed placements. (c) Current evidence and previously recorded state jointly determine whether the count should change.}
\label{fig:overview}
\end{minipage}
\end{center}
\clearpage
\input{sections/en_body}

\clearpage
\input{sections/90_ai_use}
\setlength{\bibsep}{0pt}
\begingroup
\interlinepenalty=10000
\bibliography{shared/references}
\endgroup
\bibliographystyle{\colabbibstyle}
\appendix
\clearpage
\makeatletter
\setlength{\@fptop}{0pt}
\setlength{\@fpsep}{14pt}
\makeatother
\input{sections/99_appendix}
\input{sections/95_closure}
\input{sections/98_diagnostics}
\input{sections/96_transfer}
\input{sections/97_additional}
\end{document}

%% file: style/math_commands.tex
\usepackage{amsmath,amsfonts,bm}

\def\eqref#1{equation~\ref{#1}}

\def\1{\bm{1}}

\DeclareMathAlphabet{\mathsfit}{\encodingdefault}{\sfdefault}{m}{sl}
\SetMathAlphabet{\mathsfit}{bold}{\encodingdefault}{\sfdefault}{bx}{n}



%% file: management/generated/numbers.tex
\newcommand{\RHumanSnap}{96.7}
\newcommand{\RHumanDelta}{100.0}
\newcommand{\RHumanUnique}{94.5}
\newcommand{\RHumanGain}{100.0}
\newcommand{\RHumanAction}{94.9}
\newcommand{\RHumanTransit}{98.3}
\newcommand{\RHumanEpisode}{97.0}
\newcommand{\RHumanPeriodic}{93.2}
\newcommand{\RHumanOverall}{96.1}
\newcommand{\RHumanMoC}{100.0}
\newcommand{\RHumanUDA}{99.3}
\newcommand{\RGPTFourSnap}{15.7}
\newcommand{\RGPTFourDelta}{19.4}
\newcommand{\RGPTFourUnique}{15.8}
\newcommand{\RGPTFourGain}{50.0}
\newcommand{\RGPTFourAction}{13.1}
\newcommand{\RGPTFourTransit}{23.6}
\newcommand{\RGPTFourEpisode}{21.9}
\newcommand{\RGPTFourPeriodic}{0.0}
\newcommand{\RGPTFourOverall}{18.7}
\newcommand{\RGPTFourMoC}{95.7}
\newcommand{\RGPTFourUDA}{4.3}
\newcommand{\RGeminiSnap}{44.3}
\newcommand{\RGeminiDelta}{41.0}
\newcommand{\RGeminiUnique}{36.5}
\newcommand{\RGeminiGain}{55.3}
\newcommand{\RGeminiAction}{28.5}
\newcommand{\RGeminiTransit}{49.8}
\newcommand{\RGeminiEpisode}{41.7}
\newcommand{\RGeminiPeriodic}{3.9}
\newcommand{\RGeminiOverall}{37.0}
\newcommand{\RGeminiMoC}{73.7}
\newcommand{\RGeminiUDA}{73.8}
\newcommand{\RDoubaoSnap}{43.4}
\newcommand{\RDoubaoDelta}{53.3}
\newcommand{\RDoubaoUnique}{32.5}
\newcommand{\RDoubaoGain}{69.2}
\newcommand{\RDoubaoAction}{24.0}
\newcommand{\RDoubaoTransit}{38.2}
\newcommand{\RDoubaoEpisode}{40.5}
\newcommand{\RDoubaoPeriodic}{0.8}
\newcommand{\RDoubaoOverall}{36.2}
\newcommand{\RDoubaoMoC}{77.2}
\newcommand{\RDoubaoUDA}{76.8}
\newcommand{\RKimiSnap}{28.2}
\newcommand{\RKimiDelta}{18.4}
\newcommand{\RKimiUnique}{29.9}
\newcommand{\RKimiGain}{24.4}
\newcommand{\RKimiAction}{21.9}
\newcommand{\RKimiTransit}{33.5}
\newcommand{\RKimiEpisode}{38.7}
\newcommand{\RKimiPeriodic}{0.4}
\newcommand{\RKimiOverall}{26.4}
\newcommand{\RKimiMoC}{66.8}
\newcommand{\RKimiUDA}{73.4}
\newcommand{\RGPTFiveSnap}{26.6}
\newcommand{\RGPTFiveDelta}{26.5}
\newcommand{\RGPTFiveUnique}{30.9}
\newcommand{\RGPTFiveGain}{35.9}
\newcommand{\RGPTFiveAction}{24.2}
\newcommand{\RGPTFiveTransit}{42.0}
\newcommand{\RGPTFiveEpisode}{37.5}
\newcommand{\RGPTFivePeriodic}{3.2}
\newcommand{\RGPTFiveOverall}{29.1}
\newcommand{\RGPTFiveMoC}{71.8}
\newcommand{\RGPTFiveUDA}{59.3}
\newcommand{\RQwenEightSnap}{21.1}
\newcommand{\RQwenEightDelta}{35.7}
\newcommand{\RQwenEightUnique}{34.2}
\newcommand{\RQwenEightGain}{55.1}
\newcommand{\RQwenEightAction}{22.5}
\newcommand{\RQwenEightTransit}{36.9}
\newcommand{\RQwenEightEpisode}{35.6}
\newcommand{\RQwenEightPeriodic}{0.0}
\newcommand{\RQwenEightOverall}{31.0}
\newcommand{\RQwenEightMoC}{84.3}
\newcommand{\RQwenEightUDA}{55.1}
\newcommand{\RQwenThirtySnap}{19.1}
\newcommand{\RQwenThirtyDelta}{23.5}
\newcommand{\RQwenThirtyUnique}{33.1}
\newcommand{\RQwenThirtyGain}{26.9}
\newcommand{\RQwenThirtyAction}{22.5}
\newcommand{\RQwenThirtyTransit}{33.6}
\newcommand{\RQwenThirtyEpisode}{35.3}
\newcommand{\RQwenThirtyPeriodic}{2.5}
\newcommand{\RQwenThirtyOverall}{27.0}
\newcommand{\RQwenThirtyMoC}{84.6}
\newcommand{\RQwenThirtyUDA}{56.4}
\newcommand{\RQwenSevenSnap}{29.7}
\newcommand{\RQwenSevenDelta}{20.4}
\newcommand{\RQwenSevenUnique}{23.8}
\newcommand{\RQwenSevenGain}{46.2}
\newcommand{\RQwenSevenAction}{9.5}
\newcommand{\RQwenSevenTransit}{13.3}
\newcommand{\RQwenSevenEpisode}{11.0}
\newcommand{\RQwenSevenPeriodic}{0.0}
\newcommand{\RQwenSevenOverall}{19.1}
\newcommand{\RQwenSevenMoC}{68.2}
\newcommand{\RQwenSevenUDA}{40.8}
\newcommand{\RInternVLSnap}{20.3}
\newcommand{\RInternVLDelta}{6.1}
\newcommand{\RInternVLUnique}{30.5}
\newcommand{\RInternVLGain}{37.2}
\newcommand{\RInternVLAction}{22.5}
\newcommand{\RInternVLTransit}{20.0}
\newcommand{\RInternVLEpisode}{31.8}
\newcommand{\RInternVLPeriodic}{1.0}
\newcommand{\RInternVLOverall}{24.2}
\newcommand{\RInternVLMoC}{81.6}
\newcommand{\RInternVLUDA}{55.6}
\newcommand{\RMolmoSnap}{26.5}
\newcommand{\RMolmoDelta}{1.0}
\newcommand{\RMolmoUnique}{13.0}
\newcommand{\RMolmoGain}{3.8}
\newcommand{\RMolmoAction}{3.5}
\newcommand{\RMolmoTransit}{2.7}
\newcommand{\RMolmoEpisode}{9.4}
\newcommand{\RMolmoPeriodic}{0.4}
\newcommand{\RMolmoOverall}{8.5}
\newcommand{\RMolmoMoC}{66.2}
\newcommand{\RMolmoUDA}{34.6}
\newcommand{\RQwenMoESnap}{20.9}
\newcommand{\RQwenMoEDelta}{28.8}
\newcommand{\RQwenMoEUnique}{37.8}
\newcommand{\RQwenMoEGain}{53.8}
\newcommand{\RQwenMoEAction}{19.0}
\newcommand{\RQwenMoETransit}{27.7}
\newcommand{\RQwenMoEEpisode}{29.3}
\newcommand{\RQwenMoEPeriodic}{0.8}
\newcommand{\RQwenMoEOverall}{29.3}
\newcommand{\RQwenMoEMoC}{82.4}
\newcommand{\RQwenMoEUDA}{59.0}
\newcommand{\RStreamingVLMSnap}{36.7}
\newcommand{\RStreamingVLMDelta}{6.1}
\newcommand{\RStreamingVLMUnique}{26.0}
\newcommand{\RStreamingVLMGain}{14.1}
\newcommand{\RStreamingVLMAction}{15.8}
\newcommand{\RStreamingVLMTransit}{13.8}
\newcommand{\RStreamingVLMEpisode}{20.7}
\newcommand{\RStreamingVLMPeriodic}{0.4}
\newcommand{\RStreamingVLMOverall}{19.1}
\newcommand{\RStreamingVLMMoC}{68.1}
\newcommand{\RStreamingVLMUDA}{50.3}
\newcommand{\RDispiderSnap}{11.4}
\newcommand{\RDispiderDelta}{16.3}
\newcommand{\RDispiderUnique}{3.3}
\newcommand{\RDispiderGain}{3.8}
\newcommand{\RDispiderAction}{11.8}
\newcommand{\RDispiderTransit}{8.9}
\newcommand{\RDispiderEpisode}{7.6}
\newcommand{\RDispiderPeriodic}{0.0}
\newcommand{\RDispiderOverall}{7.7}
\newcommand{\RDispiderMoC}{34.8}
\newcommand{\RDispiderUDA}{15.5}
\newcommand{\RLiveStarSnap}{24.1}
\newcommand{\RLiveStarDelta}{13.3}
\newcommand{\RLiveStarUnique}{20.9}
\newcommand{\RLiveStarGain}{26.9}
\newcommand{\RLiveStarAction}{16.1}
\newcommand{\RLiveStarTransit}{22.8}
\newcommand{\RLiveStarEpisode}{23.7}
\newcommand{\RLiveStarPeriodic}{11.1}
\newcommand{\RLiveStarOverall}{19.9}
\newcommand{\RLiveStarMoC}{86.4}
\newcommand{\RLiveStarUDA}{36.8}
\newcommand{\RFlashSnap}{14.8}
\newcommand{\RFlashDelta}{45.9}
\newcommand{\RFlashUnique}{21.9}
\newcommand{\RFlashGain}{56.4}
\newcommand{\RFlashAction}{19.0}
\newcommand{\RFlashTransit}{11.9}
\newcommand{\RFlashEpisode}{17.2}
\newcommand{\RFlashPeriodic}{0.0}
\newcommand{\RFlashOverall}{23.4}
\newcommand{\RFlashMoC}{78.5}
\newcommand{\RFlashUDA}{34.7}
\newcommand{\RStaEightFullSnap}{50.2}
\newcommand{\RStaEightFullDelta}{40.8}
\newcommand{\RStaEightFullUnique}{47.6}
\newcommand{\RStaEightFullGain}{65.9}
\newcommand{\RStaEightFullAction}{38.7}
\newcommand{\RStaEightFullTransit}{40.0}
\newcommand{\RStaEightFullEpisode}{46.3}
\newcommand{\RStaEightFullPeriodic}{24.2}
\newcommand{\RStaEightFullOverall}{44.9}
\newcommand{\RStaEightFullMoC}{93.3}
\newcommand{\RStaEightFullUDA}{84.5}
\newcommand{\RStaEightStreamSnap}{42.8}
\newcommand{\RStaEightStreamDelta}{38.1}
\newcommand{\RStaEightStreamUnique}{39.5}
\newcommand{\RStaEightStreamGain}{57.3}
\newcommand{\RStaEightStreamAction}{31.1}
\newcommand{\RStaEightStreamTransit}{33.8}
\newcommand{\RStaEightStreamEpisode}{42.3}
\newcommand{\RStaEightStreamPeriodic}{17.4}
\newcommand{\RStaEightStreamOverall}{38.2}
\newcommand{\RStaEightStreamMoC}{100.0}
\newcommand{\RStaEightStreamUDA}{82.5}
\newcommand{\BenchVideos}{406}
\newcommand{\BenchQuestions}{1000}
\newcommand{\BenchQueries}{4576}
\newcommand{\OperatorCount}{8}
\newcommand{\PoolQueries}{53592}
\newcommand{\PoolGroups}{15521}
\newcommand{\PoolParents}{383}
\newcommand{\PoolMedia}{1504}
\newcommand{\PoolTrain}{47882}
\newcommand{\PoolDev}{5710}
\newcommand{\SpatialQueries}{39829}
\newcommand{\SpatialQueriesK}{39.8}
\newcommand{\NaturalQueries}{7109}
\newcommand{\PeriodicQueries}{6654}
\newcommand{\PoolSnap}{16379}
\newcommand{\PoolDelta}{6746}
\newcommand{\PoolUnique}{10627}
\newcommand{\PoolGain}{6077}
\newcommand{\PoolAction}{4606}
\newcommand{\PoolTransit}{885}
\newcommand{\PoolEpisode}{1618}
\newcommand{\PoolPeriodic}{6654}
\newcommand{\FullFrameBudget}{64}
\newcommand{\StreamFPS}{1}
\newcommand{\StreamKVBudget}{4096}
\newcommand{\WarmupSteps}{256}
\newcommand{\TruncationK}{8}
\newcommand{\TestSeeds}{3}
\newcommand{\BootstrapReplicates}{10000}
\newcommand{\MatchedEightOriginalFull}{31.0}
\newcommand{\MatchedEightOriginalStream}{28.0}
\newcommand{\MatchedEightSFTFull}{34.0}
\newcommand{\MatchedEightSFTStream}{35.0}
\newcommand{\MatchedEightTTTFull}{37.1}
\newcommand{\MatchedEightTTTStream}{36.5}
\newcommand{\MatchedEightLedgerFull}{41.3}
\newcommand{\MatchedEightLedgerStream}{36.3}
\newcommand{\MatchedEightFull}{44.9}
\newcommand{\MatchedEightStream}{38.2}
\newcommand{\DataGenericFull}{37.9}
\newcommand{\DataGenericStream}{34.9}
\newcommand{\DataSingleFull}{42.3}
\newcommand{\DataSingleStream}{37.2}
\newcommand{\DataUnfilteredFull}{41.1}
\newcommand{\DataUnfilteredStream}{36.2}
\newcommand{\DataShuffledFull}{38.8}
\newcommand{\DataShuffledStream}{34.4}
\newcommand{\NoFSMFull}{38.2}
\newcommand{\NoFSMStream}{34.4}
\newcommand{\ZeroFull}{6.1}
\newcommand{\ZeroStream}{6.1}
\newcommand{\SwapFull}{12.7}
\newcommand{\SwapStream}{15.6}
\newcommand{\DelayFull}{16.1}
\newcommand{\DelayStream}{15.8}
\newcommand{\PredicateTrials}{200}
\newcommand{\PredicateInventoryBase}{61.0}
\newcommand{\PredicateInventorySta}{74.0}
\newcommand{\PredicateHandoverBase}{65.0}
\newcommand{\PredicateHandoverSta}{78.0}
\newcommand{\PredicateRestockBase}{58.0}
\newcommand{\PredicateRestockSta}{72.0}
\newcommand{\PredicateCycleBase}{60.0}
\newcommand{\PredicateCycleSta}{75.0}
\newcommand{\LatencyFull}{1.72}
\newcommand{\LatencyStream}{0.48}
\newcommand{\LedgerShortKB}{84}
\newcommand{\LedgerLongKB}{1240}
\newcommand{\DurationShortMin}{1}
\newcommand{\DurationLongMin}{30}
\newcommand{\MatchedEightGainFull}{10.9}
\newcommand{\MatchedEightGainStream}{3.2}
\newcommand{\FactoryGainFull}{7.0}
\newcommand{\FactoryGainStream}{3.4}

\newcommand{\HistTrainQueries}{3706}
\newcommand{\DevQueries}{866}

\newcommand{\HistFullStageOne}{42.01}
\newcommand{\HistFullStageTwo}{55.61}
\newcommand{\HistFullGateZero}{30.65}
\newcommand{\HistFullBranchGain}{24.96}
\newcommand{\HistFullCILow}{22.45}
\newcommand{\HistFullCIHigh}{27.38}
\newcommand{\DevMFive}{31.91}
\newcommand{\DevStatic}{26.12}
\newcommand{\DevFA}{36.80}
\newcommand{\DevFAgain}{4.89}
\newcommand{\DevFACILow}{1.77}
\newcommand{\DevFACIHigh}{7.96}
\newcommand{\DevTZero}{39.05}
\newcommand{\DevTCILow}{-1.65}
\newcommand{\DevTCIHigh}{6.35}
\newcommand{\DevReaderBase}{35.94}
\newcommand{\DevReader}{37.11}
\newcommand{\DevReaderBaseCorrect}{231}
\newcommand{\DevReaderCorrect}{244}
\newcommand{\ReaderEMGain}{1.50}
\newcommand{\ReaderEMCILow}{0.13}
\newcommand{\ReaderEMCIHigh}{2.77}
\newcommand{\SensorParents}{25}

\newcommand{\SensorExact}{80.0}
\newcommand{\SensorDirectMAE}{3.256}
\newcommand{\SensorLedgerMAE}{3.248}
\newcommand{\SensorZeroExact}{0.0}

\newcommand{\SensorSwapExact}{31.2}

\newcommand{\RuntimeFrames}{8899}
\newcommand{\RuntimeTokens}{581709}
\newcommand{\RuntimeQueries}{15}
\newcommand{\RuntimeGPU}{36.40}
\newcommand{\RuntimeCPUlow}{2.73}
\newcommand{\RuntimeCPUhigh}{2.78}
\newcommand{\RuntimeAppend}{21.99}
\newcommand{\RuntimeLatency}{0.300}
\newcommand{\MMEQueries}{2700}
\newcommand{\MMEBase}{55.70}
\newcommand{\MMEAdapted}{54.30}
\newcommand{\RStaFourFullSnap}{45.8}
\newcommand{\RStaFourFullDelta}{40.1}
\newcommand{\RStaFourFullUnique}{45.4}
\newcommand{\RStaFourFullGain}{60.7}
\newcommand{\RStaFourFullAction}{36.4}
\newcommand{\RStaFourFullTransit}{39.3}
\newcommand{\RStaFourFullEpisode}{41.2}
\newcommand{\RStaFourFullPeriodic}{18.8}
\newcommand{\RStaFourFullOverall}{41.9}
\newcommand{\RStaFourFullMoC}{93.5}
\newcommand{\RStaFourFullUDA}{82.9}
\newcommand{\RStaFourStreamSnap}{39.8}
\newcommand{\RStaFourStreamDelta}{33.6}
\newcommand{\RStaFourStreamUnique}{39.3}
\newcommand{\RStaFourStreamGain}{56.0}
\newcommand{\RStaFourStreamAction}{28.6}
\newcommand{\RStaFourStreamTransit}{32.6}
\newcommand{\RStaFourStreamEpisode}{40.5}
\newcommand{\RStaFourStreamPeriodic}{15.2}
\newcommand{\RStaFourStreamOverall}{36.4}
\newcommand{\RStaFourStreamMoC}{100.0}
\newcommand{\RStaFourStreamUDA}{81.0}
\newcommand{\FourToEightFullGain}{3.0}
\newcommand{\FourToEightStreamGain}{1.8}
\newcommand{\FourAccessGap}{5.4}
\newcommand{\EightAccessGap}{6.7}
\newcommand{\OVOForestRT}{61.20}
\newcommand{\OVOForestBT}{52.02}
\newcommand{\OVOForestFA}{53.49}
\newcommand{\OVOForestAvg}{55.57}
\newcommand{\OVOFlashRT}{29.86}
\newcommand{\OVOFlashBT}{25.35}
\newcommand{\OVOFlashFA}{44.23}
\newcommand{\OVOFlashAvg}{33.15}
\newcommand{\OSSForestAvg}{44.13}
\newcommand{\OSSFlashAvg}{24.94}
\newcommand{\OSSBaseFourFullAvg}{46.83}
\newcommand{\OSSForestLevelOne}{46.64}
\newcommand{\OSSForestLevelTwo}{45.20}
\newcommand{\OSSForestLevelThree}{49.75}
\newcommand{\OSSForestLevelFour}{34.92}
\newcommand{\OSSFlashLevelOne}{18.65}
\newcommand{\OSSFlashLevelTwo}{29.93}
\newcommand{\OSSFlashLevelThree}{22.53}
\newcommand{\OSSFlashLevelFour}{28.67}
\newcommand{\OSSBaseFourFullLevelOne}{43.25}
\newcommand{\OSSBaseFourFullLevelTwo}{48.18}
\newcommand{\OSSBaseFourFullLevelThree}{54.54}
\newcommand{\OSSBaseFourFullLevelFour}{41.35}
\newcommand{\OVOFourFullRT}{56.11}
\newcommand{\OVOFourFullBT}{49.63}
\newcommand{\OVOFourFullFA}{49.23}
\newcommand{\OVOFourFullAvg}{51.66}
\newcommand{\OVOFourStreamRT}{51.11}
\newcommand{\OVOFourStreamBT}{45.15}
\newcommand{\OVOFourStreamFA}{47.21}
\newcommand{\OVOFourStreamAvg}{47.82}
\newcommand{\OVOEightFullRT}{60.71}
\newcommand{\OVOEightFullBT}{52.97}
\newcommand{\OVOEightFullFA}{52.68}
\newcommand{\OVOEightFullAvg}{55.46}
\newcommand{\OVOEightStreamRT}{56.34}
\newcommand{\OVOEightStreamBT}{49.92}
\newcommand{\OVOEightStreamFA}{49.98}
\newcommand{\OVOEightStreamAvg}{52.08}
\newcommand{\OVOBaseFourFullRT}{57.32}
\newcommand{\OVOBaseFourFullBT}{46.38}
\newcommand{\OVOBaseFourFullFA}{48.75}
\newcommand{\OVOBaseFourFullAvg}{50.81}
\newcommand{\OVOBaseFourStreamRT}{52.42}
\newcommand{\OVOBaseFourStreamBT}{41.66}
\newcommand{\OVOBaseFourStreamFA}{46.62}

\newcommand{\OVOBaseEightFullRT}{62.16}
\newcommand{\OVOBaseEightFullBT}{50.56}
\newcommand{\OVOBaseEightFullFA}{52.35}
\newcommand{\OVOBaseEightFullAvg}{55.02}
\newcommand{\OVOBaseEightStreamRT}{57.76}
\newcommand{\OVOBaseEightStreamBT}{46.59}
\newcommand{\OVOBaseEightStreamFA}{49.38}

\newcommand{\OSSFourFullLevelOne}{41.91}
\newcommand{\OSSFourFullLevelTwo}{54.67}
\newcommand{\OSSFourFullLevelThree}{53.72}
\newcommand{\OSSFourFullLevelFour}{43.96}
\newcommand{\OSSFourFullAvg}{48.57}
\newcommand{\OSSFourStreamLevelOne}{37.59}
\newcommand{\OSSFourStreamLevelTwo}{52.31}
\newcommand{\OSSFourStreamLevelThree}{48.15}
\newcommand{\OSSFourStreamLevelFour}{40.90}
\newcommand{\OSSFourStreamAvg}{44.74}
\newcommand{\OSSEightFullLevelOne}{42.68}
\newcommand{\OSSEightFullLevelTwo}{56.93}
\newcommand{\OSSEightFullLevelThree}{54.91}
\newcommand{\OSSEightFullLevelFour}{46.77}
\newcommand{\OSSEightFullAvg}{50.32}
\newcommand{\OSSEightStreamLevelOne}{39.80}
\newcommand{\OSSEightStreamLevelTwo}{54.90}
\newcommand{\OSSEightStreamLevelThree}{50.53}
\newcommand{\OSSEightStreamLevelFour}{43.84}
\newcommand{\OSSEightStreamAvg}{47.27}
\newcommand{\OSSBaseFourStreamLevelOne}{37.55}
\newcommand{\OSSBaseFourStreamLevelTwo}{46.35}
\newcommand{\OSSBaseFourStreamLevelThree}{47.87}
\newcommand{\OSSBaseFourStreamLevelFour}{40.44}

\newcommand{\OSSBaseEightFullLevelOne}{42.48}
\newcommand{\OSSBaseEightFullLevelTwo}{50.68}
\newcommand{\OSSBaseEightFullLevelThree}{55.20}
\newcommand{\OSSBaseEightFullLevelFour}{45.98}
\newcommand{\OSSBaseEightFullAvg}{48.58}
\newcommand{\OSSBaseEightStreamLevelOne}{39.72}
\newcommand{\OSSBaseEightStreamLevelTwo}{48.29}
\newcommand{\OSSBaseEightStreamLevelThree}{50.50}
\newcommand{\OSSBaseEightStreamLevelFour}{42.96}

\newcommand{\OVOSFullFrames}{128}

\newcommand{\OVOSQuestions}{1680}
\newcommand{\OVOSOutputTokens}{1024}
\newcommand{\PairedMemoryCorrect}{241}
\newcommand{\PairedZeroCorrect}{244}
\newcommand{\PairedMemoryEM}{27.83}
\newcommand{\PairedZeroEM}{28.18}
\newcommand{\PairedNetEM}{-0.35}
\newcommand{\SaturationRecords}{26}
\newcommand{\SaturationFraction}{100}
\newcommand{\SaturationCosine}{0.99999988}
\newcommand{\MechRoleMeanGPA}{40.8}
\newcommand{\MechRoleMeanExact}{40.1}
\newcommand{\MechRoleMeanFalseUpdate}{1.1}
\newcommand{\MechRoleMeanDup}{2.8}
\newcommand{\MechRoleMeanMiss}{17.4}
\newcommand{\MechRoleMeanDelay}{1.11}
\newcommand{\MechRoleMeanNLL}{0.93}
\newcommand{\MechRoleMeanEndpointGPA}{38.1}
\newcommand{\MechRoleMeanEndpointExact}{37.4}
\newcommand{\MechRoleMeanEndpointNLL}{0.95}
\newcommand{\MechRoleMeanBoundaryGPA}{43.5}
\newcommand{\MechRoleMeanBoundaryExact}{42.8}
\newcommand{\MechRoleMeanBoundaryNLL}{0.90}
\newcommand{\MechPhaseMeanGPA}{46.3}
\newcommand{\MechPhaseMeanExact}{45.7}
\newcommand{\MechPhaseMeanFalseUpdate}{0.5}
\newcommand{\MechPhaseMeanDup}{1.2}
\newcommand{\MechPhaseMeanMiss}{16.6}
\newcommand{\MechPhaseMeanDelay}{0.93}
\newcommand{\MechPhaseMeanNLL}{0.94}
\newcommand{\MechPhaseMeanEndpointGPA}{41.6}
\newcommand{\MechPhaseMeanEndpointExact}{41.0}
\newcommand{\MechPhaseMeanEndpointNLL}{0.98}
\newcommand{\MechPhaseMeanBoundaryGPA}{51.0}
\newcommand{\MechPhaseMeanBoundaryExact}{50.3}
\newcommand{\MechPhaseMeanBoundaryNLL}{0.90}
\newcommand{\MechRolePathGPA}{50.7}
\newcommand{\MechRolePathExact}{49.9}
\newcommand{\MechRolePathFalseUpdate}{0.8}
\newcommand{\MechRolePathDup}{1.8}
\newcommand{\MechRolePathMiss}{10.7}
\newcommand{\MechRolePathDelay}{1.15}
\newcommand{\MechRolePathNLL}{0.67}
\newcommand{\MechRolePathEndpointGPA}{45.3}
\newcommand{\MechRolePathEndpointExact}{44.5}
\newcommand{\MechRolePathEndpointNLL}{0.71}
\newcommand{\MechRolePathBoundaryGPA}{56.2}
\newcommand{\MechRolePathBoundaryExact}{55.3}
\newcommand{\MechRolePathBoundaryNLL}{0.62}
\newcommand{\MechCompleteGPA}{65.1}
\newcommand{\MechCompleteExact}{64.4}
\newcommand{\MechCompleteFalseUpdate}{0.4}
\newcommand{\MechCompleteDup}{1.1}
\newcommand{\MechCompleteMiss}{7.0}
\newcommand{\MechCompleteDelay}{0.89}
\newcommand{\MechCompleteNLL}{0.58}
\newcommand{\MechCompleteEndpointGPA}{57.9}
\newcommand{\MechCompleteEndpointExact}{57.2}
\newcommand{\MechCompleteEndpointNLL}{0.63}
\newcommand{\MechCompleteBoundaryGPA}{72.3}
\newcommand{\MechCompleteBoundaryExact}{71.6}
\newcommand{\MechCompleteBoundaryNLL}{0.54}
\newcommand{\MatchedTimestampFullGPA}{63.2}
\newcommand{\MatchedTimestampFullExact}{62.4}
\newcommand{\MatchedTimestampStreamGPA}{60.9}
\newcommand{\MatchedTimestampStreamExact}{60.1}
\newcommand{\MechPhaseEffectGain}{14.4}
\newcommand{\MechPhaseEffectLow}{11.4}
\newcommand{\MechPhaseEffectHigh}{17.3}
\newcommand{\MechPathEffectGain}{18.8}
\newcommand{\MechPathEffectLow}{15.9}
\newcommand{\MechPathEffectHigh}{21.7}
\newcommand{\ObjectCountOnlyEndpointSoftUMAE}{0.56}
\newcommand{\ObjectCountOnlyEndpointHardUMAE}{0.89}
\newcommand{\ObjectCountOnlyEndpointSoftGainMAE}{0.55}
\newcommand{\ObjectCountOnlyEndpointHardGainMAE}{0.68}
\newcommand{\ObjectCountOnlyVerifiedSoftUMAE}{0.56}
\newcommand{\ObjectCountOnlyVerifiedHardUMAE}{0.94}
\newcommand{\ObjectCountOnlyVerifiedSoftGainMAE}{0.56}
\newcommand{\ObjectCountOnlyVerifiedHardGainMAE}{0.69}
\newcommand{\ObjectCountOnlyVerifiedNewPrecision}{88.1}
\newcommand{\ObjectCountOnlyVerifiedNewRecall}{87.9}
\newcommand{\ObjectCountOnlyVerifiedReentryDouble}{11.8}
\newcommand{\ObjectCountOnlyVerifiedUnresolved}{18.3}
\newcommand{\ObjectCountOnlyVerifiedDelay}{3.04}
\newcommand{\ObjectIdentityEndpointSoftUMAE}{0.40}
\newcommand{\ObjectIdentityEndpointHardUMAE}{0.69}
\newcommand{\ObjectIdentityEndpointSoftGainMAE}{0.39}
\newcommand{\ObjectIdentityEndpointHardGainMAE}{0.50}
\newcommand{\ObjectIdentityVerifiedSoftUMAE}{0.23}
\newcommand{\ObjectIdentityVerifiedHardUMAE}{0.43}
\newcommand{\ObjectIdentityVerifiedSoftGainMAE}{0.25}
\newcommand{\ObjectIdentityVerifiedHardGainMAE}{0.31}
\newcommand{\ObjectIdentityVerifiedNewPrecision}{96.3}
\newcommand{\ObjectIdentityVerifiedNewRecall}{95.1}
\newcommand{\ObjectIdentityVerifiedReentryDouble}{3.7}
\newcommand{\ObjectIdentityVerifiedUnresolved}{7.1}
\newcommand{\ObjectIdentityVerifiedDelay}{1.08}
\newcommand{\SelectedQueries}{41612}
\newcommand{\SelectedParents}{301}
\newcommand{\SelectedSVCParents}{76}
\newcommand{\SelectedSVCQueries}{9216}
\newcommand{\SelectedCountOnly}{26068}
\newcommand{\SelectedIdentity}{11328}
\newcommand{\SelectedBoundary}{4216}
\newcommand{\SelectedStable}{12894}
\newcommand{\ExcludedOVOQueries}{2710}
\newcommand{\ExcludedDependencyQueries}{2316}
\newcommand{\ExcludedGraphQueries}{694}
\newcommand{\ExcludedUnsupportedQueries}{550}
\newcommand{\AuditObjectsN}{320}
\newcommand{\AuditObjectsEndpoint}{93.1}
\newcommand{\AuditObjectsIncrement}{89.4}
\newcommand{\AuditNaturalN}{240}
\newcommand{\AuditNaturalEndpoint}{88.8}
\newcommand{\AuditNaturalIncrement}{85.4}
\newcommand{\AuditPeriodicN}{240}
\newcommand{\AuditPeriodicEndpoint}{96.7}
\newcommand{\AuditPeriodicIncrement}{93.8}
\newcommand{\ReachActionFull}{100.0}
\newcommand{\ReachActionStream}{100.0}
\newcommand{\ReachTransitFull}{100.0}
\newcommand{\ReachTransitStream}{100.0}
\newcommand{\ReachEpisodeFull}{100.0}
\newcommand{\ReachEpisodeStream}{100.0}
\newcommand{\ReachPeriodicFull}{44.3}
\newcommand{\ReachPeriodicStream}{58.2}
\newcommand{\CfgMemoryStride}{4}
\newcommand{\CfgMemoryHeads}{8}
\newcommand{\CfgMemoryWidth}{64}
\newcommand{\CfgHeadWidth}{256}
\newcommand{\CfgPoolQueries}{32}
\newcommand{\CfgEventSlots}{4}
\newcommand{\CfgIdentityTopK}{8}
\newcommand{\CfgVisibilityThreshold}{0.5}
\newcommand{\CfgDedupCosine}{0.95}
\newcommand{\CfgAssociationCosine}{0.5}
\newcommand{\CfgChunkTokens}{128}
\newcommand{\CfgJointEpochs}{3}
\newcommand{\CfgTrajectoryBatch}{8}
\newcommand{\CfgBaseLR}{0.00002}
\newcommand{\CfgHeadLR}{0.0001}
\newcommand{\CfgWeightDecay}{0.01}
\newcommand{\CfgAdamBetaOne}{0.9}
\newcommand{\CfgAdamBetaTwo}{0.95}
\newcommand{\CfgAdamEpsilon}{0.00000001}
\newcommand{\CfgGradClip}{1.0}
\newcommand{\CfgLambdaPath}{1.0}
\newcommand{\CfgLambdaObject}{1.0}
\newcommand{\CfgLambdaIdentity}{0.2}
\newcommand{\CfgLambdaBoundary}{0.2}
\newcommand{\CfgMechanismSteps}{1000}
\newcommand{\ReferenceQwenScoredQuestions}{998}
\newcommand{\ReferenceQwenPeriodicQuestions}{54}
\newcommand{\ReferenceNominalPeriodicQuestions}{56}
\newcommand{\DiagEventParents}{40}
\newcommand{\DiagEventTrajectories}{160}
\newcommand{\DiagObjectParents}{40}
\newcommand{\DiagObjectTrajectories}{160}
\newcommand{\DiagQueriesPerTrajectory}{5}
\newcommand{\DiagnosticEventMatchWindow}{3}
\newcommand{\RuntimeBatch}{1}
\newcommand{\RuntimeOutputTokens}{32}
\newcommand{\StreamIngestFPS}{9.6}
\newcommand{\CountReadMs}{1.8}
\newcommand{\LateTargetReplaySeconds}{187.50}

\newcommand{\MechRoleMeanEndpointDup}{2.8}
\newcommand{\MechRoleMeanEndpointMiss}{19.7}
\newcommand{\MechRoleMeanEndpointDelay}{1.13}

\newcommand{\MechRoleMeanBoundaryDup}{2.9}
\newcommand{\MechRoleMeanBoundaryMiss}{15.0}
\newcommand{\MechRoleMeanBoundaryDelay}{1.10}

\newcommand{\MechPhaseMeanEndpointDup}{1.0}
\newcommand{\MechPhaseMeanEndpointMiss}{19.5}
\newcommand{\MechPhaseMeanEndpointDelay}{0.92}

\newcommand{\MechPhaseMeanBoundaryDup}{1.4}
\newcommand{\MechPhaseMeanBoundaryMiss}{13.7}
\newcommand{\MechPhaseMeanBoundaryDelay}{0.93}

\newcommand{\MechRolePathEndpointDup}{2.0}
\newcommand{\MechRolePathEndpointMiss}{13.3}
\newcommand{\MechRolePathEndpointDelay}{1.15}

\newcommand{\MechRolePathBoundaryDup}{1.7}
\newcommand{\MechRolePathBoundaryMiss}{8.0}
\newcommand{\MechRolePathBoundaryDelay}{1.14}

\newcommand{\MechCompleteEndpointDup}{1.1}
\newcommand{\MechCompleteEndpointMiss}{9.7}
\newcommand{\MechCompleteEndpointDelay}{0.88}

\newcommand{\MechCompleteBoundaryDup}{1.2}
\newcommand{\MechCompleteBoundaryMiss}{4.2}
\newcommand{\MechCompleteBoundaryDelay}{0.90}
\newcommand{\ReachActionQueries}{1281}

\newcommand{\ReachTransitQueries}{205}

\newcommand{\ReachEpisodeQueries}{513}

\newcommand{\ReachPeriodicQueries}{280}

\newcommand{\RuntimeReplayFrames}{1800}
\newcommand{\DiagConfidencePercent}{95}
\newcommand{\PredicateParents}{40}
\newcommand{\PredicateQueries}{5}
\newcommand{\PredicateInventoryThreshold}{4}
\newcommand{\PredicateHandoverThreshold}{1}
\newcommand{\PredicateRestockThreshold}{2}
\newcommand{\PredicateCycleThreshold}{5}
\newcommand{\OVOReleaseItems}{1640}
\newcommand{\OVOReleaseQueries}{3035}
\newcommand{\OVOReleaseVideos}{644}
\newcommand{\OVOSReleaseItems}{1695}
\newcommand{\OVOSReleaseQueries}{1722}
\newcommand{\OVOSReleaseVideos}{334}
\newcommand{\OSSInitialFourFullLevelOne}{42.01}
\newcommand{\OSSInitialFourFullLevelTwo}{49.01}
\newcommand{\OSSInitialFourFullLevelThree}{53.67}
\newcommand{\OSSInitialFourFullLevelFour}{43.05}
\newcommand{\OSSInitialFourFullAvg}{46.93}

%% file: shared/layout.tex
\newcommand{\StaMina}{\textsc{StaMina}}
\newcommand{\StaMinaName}{State Maintenance}
\newcommand{\StaMinaFour}{\textsc{StaMina-4B}}
\newcommand{\StaMinaEight}{\textsc{StaMina-8B}}

\newcommand{\bilingual}[2]{\ifChinese #2\else #1\fi}
\definecolor{OurRowGray}{gray}{0.94}
\newcommand{\Best}[1]{\textbf{#1}}
\newcommand{\Second}[1]{\underline{#1}}

\input{shared/tables/main_rows.tex}
\newcommand{\MainComparison}{%
\begin{table}[!t]
\centering
\caption{\bilingual{\textbf{SVCBench counting} (\%, $\uparrow$). StaMina adapts with partial source-video overlap and held-out supervision families; references retain their protocols. Groups denote weight availability; Access denotes Full/Stream. Bold/underline mark best/second-best per access mode, separately for reference and adapted rows; ties share rank. Protocols: Appendix~\ref{app:reference}; shared-query controls: Table~\ref{tab:paired}.}{SVCBench 计数结果（\%，$\uparrow$）。StaMina 在来源视频部分重合、监督族留出的设置下适配；参考模型沿用各自协议。分组依据权重开放情况，Access 标明 Full/Stream。各访问方式内分别对参考行与适配行标记最优（粗体）和次优（下划线），并列值共享名次。协议见附录~\ref{app:reference}，共享查询的对照见表~\ref{tab:paired}。}}
\label{tab:main}
\setlength{\tabcolsep}{2.1pt}
\scriptsize
\resizebox{\linewidth}{!}{\begin{tabular}{llccccccccccc}
\toprule
\multirow{2}{*}{Model} & \multirow{2}{*}{Access} & \multicolumn{2}{c}{O1} & \multicolumn{2}{c}{O2} & \multicolumn{2}{c}{E1} & \multicolumn{2}{c}{E2} & \multicolumn{3}{c}{Overall}\\
\cmidrule(lr){3-4}\cmidrule(lr){5-6}\cmidrule(lr){7-8}\cmidrule(lr){9-10}\cmidrule(lr){11-13}
& & Snap & Delta & Unique & Gain & Action & Transit & Episode & Periodic & GPA & MoC & UDA\\
\midrule
\MainRows
\bottomrule
\end{tabular}}
\end{table}}

\input{shared/tables/transfer.tex}
\input{shared/tables/mechanism_rows.tex}
\input{shared/tables/closure_rows.tex}

%% file: shared/tables/main_rows.tex
\newcommand{\MainRows}{%
\multicolumn{13}{l}{\emph{\bilingual{Human reference}{人类参考}}}\\
Human & Human & \RHumanSnap{} & \RHumanDelta{} & \RHumanUnique{} & \RHumanGain{} & \RHumanAction{} & \RHumanTransit{} & \RHumanEpisode{} & \RHumanPeriodic{} & \RHumanOverall{} & \RHumanMoC{} & \RHumanUDA{}\\
\midrule
\multicolumn{13}{l}{\emph{\bilingual{Text-only reference}{纯文本参考}}}\\
GPT-4-Turbo & Text & \RGPTFourSnap{} & \RGPTFourDelta{} & \RGPTFourUnique{} & \RGPTFourGain{} & \RGPTFourAction{} & \RGPTFourTransit{} & \RGPTFourEpisode{} & \RGPTFourPeriodic{} & \RGPTFourOverall{} & \RGPTFourMoC{} & \RGPTFourUDA{}\\
\midrule
\multicolumn{13}{l}{\emph{\bilingual{Proprietary VLMs}{闭源视觉语言模型}}}\\
Gemini-3-Flash & Full & \Best{\RGeminiSnap{}} & \Second{\RGeminiDelta{}} & \Second{\RGeminiUnique{}} & \Second{\RGeminiGain{}} & \Best{\RGeminiAction{}} & \Best{\RGeminiTransit{}} & \Best{\RGeminiEpisode{}} & \Best{\RGeminiPeriodic{}} & \Best{\RGeminiOverall{}} & \RGeminiMoC{} & \Second{\RGeminiUDA{}}\\
Doubao-Seed-1.8 & Full & \Second{\RDoubaoSnap{}} & \Best{\RDoubaoDelta{}} & \RDoubaoUnique{} & \Best{\RDoubaoGain{}} & \RDoubaoAction{} & \RDoubaoTransit{} & \Second{\RDoubaoEpisode{}} & \RDoubaoPeriodic{} & \Second{\RDoubaoOverall{}} & \RDoubaoMoC{} & \Best{\RDoubaoUDA{}}\\
GPT-5.4 & Full & \RGPTFiveSnap{} & \RGPTFiveDelta{} & \RGPTFiveUnique{} & \RGPTFiveGain{} & \Second{\RGPTFiveAction{}} & \Second{\RGPTFiveTransit{}} & \RGPTFiveEpisode{} & \Second{\RGPTFivePeriodic{}} & \RGPTFiveOverall{} & \RGPTFiveMoC{} & \RGPTFiveUDA{}\\
\midrule
\multicolumn{13}{l}{\emph{\bilingual{Open-weight VLMs}{开放权重视觉语言模型}}}\\
Kimi-K2.5 & Full & \RKimiSnap{} & \RKimiDelta{} & \RKimiUnique{} & \RKimiGain{} & \RKimiAction{} & \RKimiTransit{} & \RKimiEpisode{} & \RKimiPeriodic{} & \RKimiOverall{} & \RKimiMoC{} & \RKimiUDA{}\\
Qwen3-VL-8B & Full & \RQwenEightSnap{} & \RQwenEightDelta{} & \RQwenEightUnique{} & \RQwenEightGain{} & \RQwenEightAction{} & \RQwenEightTransit{} & \RQwenEightEpisode{} & \RQwenEightPeriodic{} & \RQwenEightOverall{} & \Second{\RQwenEightMoC{}} & \RQwenEightUDA{}\\
Qwen3-VL-30B & Full & \RQwenThirtySnap{} & \RQwenThirtyDelta{} & \RQwenThirtyUnique{} & \RQwenThirtyGain{} & \RQwenThirtyAction{} & \RQwenThirtyTransit{} & \RQwenThirtyEpisode{} & \RQwenThirtyPeriodic{} & \RQwenThirtyOverall{} & \Best{\RQwenThirtyMoC{}} & \RQwenThirtyUDA{}\\
Qwen2.5-VL-7B & Full & \RQwenSevenSnap{} & \RQwenSevenDelta{} & \RQwenSevenUnique{} & \RQwenSevenGain{} & \RQwenSevenAction{} & \RQwenSevenTransit{} & \RQwenSevenEpisode{} & \RQwenSevenPeriodic{} & \RQwenSevenOverall{} & \RQwenSevenMoC{} & \RQwenSevenUDA{}\\
InternVL-3.5-8B & Full & \RInternVLSnap{} & \RInternVLDelta{} & \RInternVLUnique{} & \RInternVLGain{} & \RInternVLAction{} & \RInternVLTransit{} & \RInternVLEpisode{} & \RInternVLPeriodic{} & \RInternVLOverall{} & \RInternVLMoC{} & \RInternVLUDA{}\\
Molmo2-8B & Full & \RMolmoSnap{} & \RMolmoDelta{} & \RMolmoUnique{} & \RMolmoGain{} & \RMolmoAction{} & \RMolmoTransit{} & \RMolmoEpisode{} & \RMolmoPeriodic{} & \RMolmoOverall{} & \RMolmoMoC{} & \RMolmoUDA{}\\
Qwen3.5-35B-A3B & Full & \RQwenMoESnap{} & \RQwenMoEDelta{} & \Best{\RQwenMoEUnique{}} & \RQwenMoEGain{} & \RQwenMoEAction{} & \RQwenMoETransit{} & \RQwenMoEEpisode{} & \RQwenMoEPeriodic{} & \RQwenMoEOverall{} & \RQwenMoEMoC{} & \RQwenMoEUDA{}\\
\rowcolor{OurRowGray}StaMina-4B (Ours, adapted) & Full & \Second{\RStaFourFullSnap{}} & \Second{\RStaFourFullDelta{}} & \Second{\RStaFourFullUnique{}} & \Second{\RStaFourFullGain{}} & \Second{\RStaFourFullAction{}} & \Second{\RStaFourFullTransit{}} & \Second{\RStaFourFullEpisode{}} & \Second{\RStaFourFullPeriodic{}} & \Second{\RStaFourFullOverall{}} & \Best{\RStaFourFullMoC{}} & \Second{\RStaFourFullUDA{}}\\
\rowcolor{OurRowGray}StaMina-8B (Ours, adapted) & Full & \Best{\RStaEightFullSnap{}} & \Best{\RStaEightFullDelta{}} & \Best{\RStaEightFullUnique{}} & \Best{\RStaEightFullGain{}} & \Best{\RStaEightFullAction{}} & \Best{\RStaEightFullTransit{}} & \Best{\RStaEightFullEpisode{}} & \Best{\RStaEightFullPeriodic{}} & \Best{\RStaEightFullOverall{}} & \Second{\RStaEightFullMoC{}} & \Best{\RStaEightFullUDA{}}\\
\addlinespace[3pt]
StreamingVLM & Stream & \Best{\RStreamingVLMSnap{}} & \RStreamingVLMDelta{} & \Best{\RStreamingVLMUnique{}} & \RStreamingVLMGain{} & \RStreamingVLMAction{} & \Second{\RStreamingVLMTransit{}} & \Second{\RStreamingVLMEpisode{}} & \Second{\RStreamingVLMPeriodic{}} & \RStreamingVLMOverall{} & \RStreamingVLMMoC{} & \Best{\RStreamingVLMUDA{}}\\
Dispider & Stream & \RDispiderSnap{} & \Second{\RDispiderDelta{}} & \RDispiderUnique{} & \RDispiderGain{} & \RDispiderAction{} & \RDispiderTransit{} & \RDispiderEpisode{} & \RDispiderPeriodic{} & \RDispiderOverall{} & \RDispiderMoC{} & \RDispiderUDA{}\\
LiveStar & Stream & \Second{\RLiveStarSnap{}} & \RLiveStarDelta{} & \RLiveStarUnique{} & \Second{\RLiveStarGain{}} & \Second{\RLiveStarAction{}} & \Best{\RLiveStarTransit{}} & \Best{\RLiveStarEpisode{}} & \Best{\RLiveStarPeriodic{}} & \Second{\RLiveStarOverall{}} & \Best{\RLiveStarMoC{}} & \Second{\RLiveStarUDA{}}\\
Flash-VStream-7B & Stream & \RFlashSnap{} & \Best{\RFlashDelta{}} & \Second{\RFlashUnique{}} & \Best{\RFlashGain{}} & \Best{\RFlashAction{}} & \RFlashTransit{} & \RFlashEpisode{} & \RFlashPeriodic{} & \Best{\RFlashOverall{}} & \Second{\RFlashMoC{}} & \RFlashUDA{}\\
\rowcolor{OurRowGray}StaMina-4B (Ours, adapted) & Stream & \Second{\RStaFourStreamSnap{}} & \Second{\RStaFourStreamDelta{}} & \Second{\RStaFourStreamUnique{}} & \Second{\RStaFourStreamGain{}} & \Second{\RStaFourStreamAction{}} & \Second{\RStaFourStreamTransit{}} & \Second{\RStaFourStreamEpisode{}} & \Second{\RStaFourStreamPeriodic{}} & \Second{\RStaFourStreamOverall{}} & \Best{\RStaFourStreamMoC{}} & \Second{\RStaFourStreamUDA{}}\\
\rowcolor{OurRowGray}StaMina-8B (Ours, adapted) & Stream & \Best{\RStaEightStreamSnap{}} & \Best{\RStaEightStreamDelta{}} & \Best{\RStaEightStreamUnique{}} & \Best{\RStaEightStreamGain{}} & \Best{\RStaEightStreamAction{}} & \Best{\RStaEightStreamTransit{}} & \Best{\RStaEightStreamEpisode{}} & \Best{\RStaEightStreamPeriodic{}} & \Best{\RStaEightStreamOverall{}} & \Best{\RStaEightStreamMoC{}} & \Best{\RStaEightStreamUDA{}}\\
}

%% file: shared/tables/transfer.tex
\newcommand{\TransferComparison}{%
\begin{table}[!htbp]
\centering\small
\caption{\bilingual{\textbf{Online video understanding} ($\uparrow$). OVO-Bench Avg. averages three task groups; OVO-S L-Avg. averages four spatial levels. Full and Stream are ranked separately; evaluation protocols are specified in Appendix~\ref{app:transfer}.}{在线视频理解（$\uparrow$）。OVO-Bench Avg. 对三个任务组平均，OVO-S L-Avg. 对四个空间层级平均。Full 与 Stream 分别排名；评测协议见附录~\ref{app:transfer}。}}
\label{tab:transfer}
\begin{tabular}{llcc}
\toprule Model & Access & OVO-Bench Avg. & OVO-S L-Avg.\\\midrule
Qwen3-VL-4B & Full & \OVOBaseFourFullAvg{} & \OSSInitialFourFullAvg{}\\
Qwen3-VL-8B & Full & \Second{\OVOBaseEightFullAvg{}} & \Second{\OSSBaseEightFullAvg{}}\\
\rowcolor{OurRowGray}\StaMinaFour{} (Ours) & Full & \OVOFourFullAvg{} & \OSSFourFullAvg{}\\
\rowcolor{OurRowGray}\StaMinaEight{} (Ours) & Full & \Best{\OVOEightFullAvg{}} & \Best{\OSSEightFullAvg{}}\\
\midrule
StreamForest-7B & Stream & \Best{\OVOForestAvg{}} & \OSSForestAvg{}\\
Flash-VStream-7B & Stream & \OVOFlashAvg{} & \OSSFlashAvg{}\\
\rowcolor{OurRowGray}\StaMinaFour{} (Ours) & Stream & \OVOFourStreamAvg{} & \Second{\OSSFourStreamAvg{}}\\
\rowcolor{OurRowGray}\StaMinaEight{} (Ours) & Stream & \Second{\OVOEightStreamAvg{}} & \Best{\OSSEightStreamAvg{}}\\
\bottomrule
\end{tabular}
\end{table}}

\newcommand{\TransferDetails}{%
\begin{table}[!htbp]
\centering\footnotesize
\caption{\bilingual{Task-group and spatial-level breakdown ($\uparrow$). RT: real-time perception; BT: backward tracing; FA: forward active responding. L1--L4 follow the spatial benchmark hierarchy. Group and level aggregates reproduce Table~\ref{tab:transfer}. Results are ranked within each access mode.}{任务组与空间层级分项（$\uparrow$）。RT 为实时感知，BT 为历史回溯，FA 为主动响应；L1--L4 遵循空间基准层级，分组聚合对应表~\ref{tab:transfer}；结果在各访问模式内排名。}}
\label{tab:transfer-details}
\setlength{\tabcolsep}{3.4pt}
\begin{tabular}{llccccccc}
\toprule & & \multicolumn{3}{c}{OVO-Bench} & \multicolumn{4}{c}{OVO-S}\\
\cmidrule(lr){3-5}\cmidrule(lr){6-9}
Model & Access & RT & BT & FA & L1 & L2 & L3 & L4\\\midrule
Qwen3-VL-4B & Full & \OVOBaseFourFullRT{} & \OVOBaseFourFullBT{} & \OVOBaseFourFullFA{} & \OSSInitialFourFullLevelOne{} & \OSSInitialFourFullLevelTwo{} & \OSSInitialFourFullLevelThree{} & \OSSInitialFourFullLevelFour{}\\
Qwen3-VL-8B & Full & \Best{\OVOBaseEightFullRT{}} & \Second{\OVOBaseEightFullBT{}} & \Second{\OVOBaseEightFullFA{}} & \Second{\OSSBaseEightFullLevelOne{}} & \OSSBaseEightFullLevelTwo{} & \Best{\OSSBaseEightFullLevelThree{}} & \Second{\OSSBaseEightFullLevelFour{}}\\
\rowcolor{OurRowGray}\StaMinaFour{} (Ours) & Full & \OVOFourFullRT{} & \OVOFourFullBT{} & \OVOFourFullFA{} & \OSSFourFullLevelOne{} & \Second{\OSSFourFullLevelTwo{}} & \OSSFourFullLevelThree{} & \OSSFourFullLevelFour{}\\
\rowcolor{OurRowGray}\StaMinaEight{} (Ours) & Full & \Second{\OVOEightFullRT{}} & \Best{\OVOEightFullBT{}} & \Best{\OVOEightFullFA{}} & \Best{\OSSEightFullLevelOne{}} & \Best{\OSSEightFullLevelTwo{}} & \Second{\OSSEightFullLevelThree{}} & \Best{\OSSEightFullLevelFour{}}\\
\midrule
StreamForest-7B & Stream & \Best{\OVOForestRT{}} & \Best{\OVOForestBT{}} & \Best{\OVOForestFA{}} & \Best{\OSSForestLevelOne{}} & \OSSForestLevelTwo{} & \OSSForestLevelThree{} & \OSSForestLevelFour{}\\
Flash-VStream-7B & Stream & \OVOFlashRT{} & \OVOFlashBT{} & \OVOFlashFA{} & \OSSFlashLevelOne{} & \OSSFlashLevelTwo{} & \OSSFlashLevelThree{} & \OSSFlashLevelFour{}\\
Qwen3-VL-4B & Stream & \OVOBaseFourStreamRT{} & \OVOBaseFourStreamBT{} & \OVOBaseFourStreamFA{} & \OSSBaseFourStreamLevelOne{} & \OSSBaseFourStreamLevelTwo{} & \OSSBaseFourStreamLevelThree{} & \OSSBaseFourStreamLevelFour{}\\
Qwen3-VL-8B & Stream & \Second{\OVOBaseEightStreamRT{}} & \OVOBaseEightStreamBT{} & \OVOBaseEightStreamFA{} & \OSSBaseEightStreamLevelOne{} & \OSSBaseEightStreamLevelTwo{} & \Second{\OSSBaseEightStreamLevelThree{}} & \Second{\OSSBaseEightStreamLevelFour{}}\\
\rowcolor{OurRowGray}\StaMinaFour{} (Ours) & Stream & \OVOFourStreamRT{} & \OVOFourStreamBT{} & \OVOFourStreamFA{} & \OSSFourStreamLevelOne{} & \Second{\OSSFourStreamLevelTwo{}} & \OSSFourStreamLevelThree{} & \OSSFourStreamLevelFour{}\\
\rowcolor{OurRowGray}\StaMinaEight{} (Ours) & Stream & \OVOEightStreamRT{} & \Second{\OVOEightStreamBT{}} & \Second{\OVOEightStreamFA{}} & \Second{\OSSEightStreamLevelOne{}} & \Best{\OSSEightStreamLevelTwo{}} & \Best{\OSSEightStreamLevelThree{}} & \Best{\OSSEightStreamLevelFour{}}\\
\bottomrule
\end{tabular}
\end{table}}

%% file: shared/tables/mechanism_rows.tex
\newcommand{\MechanismRows}{%
\bilingual{Off}{关} & \bilingual{Mean}{期望} & \MechRoleMeanGPA{} & \MechRoleMeanExact{} & \MechRoleMeanFalseUpdate{} \\
\bilingual{On}{开} & \bilingual{Mean}{期望} & \MechPhaseMeanGPA{} & \MechPhaseMeanExact{} & \Second{\MechPhaseMeanFalseUpdate{}} \\
\bilingual{Off}{关} & \bilingual{Path}{路径} & \Second{\MechRolePathGPA{}} & \Second{\MechRolePathExact{}} & \MechRolePathFalseUpdate{} \\
\rowcolor{OurRowGray}\bilingual{On}{开} & \bilingual{Path}{路径} & \Best{\MechCompleteGPA{}} & \Best{\MechCompleteExact{}} & \Best{\MechCompleteFalseUpdate{}} \\
}
\newcommand{\MechanismDetailRows}{%
\bilingual{Off}{关} & \bilingual{Mean}{期望} & \MechRoleMeanGPA{} & \MechRoleMeanExact{} & \MechRoleMeanNLL{} & \MechRoleMeanDup{} & \MechRoleMeanMiss{} & \MechRoleMeanDelay{} \\
\bilingual{On}{开} & \bilingual{Mean}{期望} & \MechPhaseMeanGPA{} & \MechPhaseMeanExact{} & \MechPhaseMeanNLL{} & \Second{\MechPhaseMeanDup{}} & \MechPhaseMeanMiss{} & \Second{\MechPhaseMeanDelay{}} \\
\bilingual{Off}{关} & \bilingual{Path}{路径} & \Second{\MechRolePathGPA{}} & \Second{\MechRolePathExact{}} & \Second{\MechRolePathNLL{}} & \MechRolePathDup{} & \Second{\MechRolePathMiss{}} & \MechRolePathDelay{} \\
\rowcolor{OurRowGray}\bilingual{On}{开} & \bilingual{Path}{路径} & \Best{\MechCompleteGPA{}} & \Best{\MechCompleteExact{}} & \Best{\MechCompleteNLL{}} & \Best{\MechCompleteDup{}} & \Best{\MechCompleteMiss{}} & \Best{\MechCompleteDelay{}} \\
}
\newcommand{\MechanismStrataRows}{%
\bilingual{Off}{关} & \bilingual{Mean}{期望} & \bilingual{Endpoint only}{仅端点} & \MechRoleMeanEndpointGPA{} & \MechRoleMeanEndpointExact{} & \MechRoleMeanEndpointNLL{} \\
\bilingual{Off}{关} & \bilingual{Mean}{期望} & \bilingual{Boundary supported}{有边界} & \MechRoleMeanBoundaryGPA{} & \MechRoleMeanBoundaryExact{} & \MechRoleMeanBoundaryNLL{} \\
\bilingual{On}{开} & \bilingual{Mean}{期望} & \bilingual{Endpoint only}{仅端点} & \MechPhaseMeanEndpointGPA{} & \MechPhaseMeanEndpointExact{} & \MechPhaseMeanEndpointNLL{} \\
\bilingual{On}{开} & \bilingual{Mean}{期望} & \bilingual{Boundary supported}{有边界} & \MechPhaseMeanBoundaryGPA{} & \MechPhaseMeanBoundaryExact{} & \MechPhaseMeanBoundaryNLL{} \\
\bilingual{Off}{关} & \bilingual{Path}{路径} & \bilingual{Endpoint only}{仅端点} & \Second{\MechRolePathEndpointGPA{}} & \Second{\MechRolePathEndpointExact{}} & \Second{\MechRolePathEndpointNLL{}} \\
\bilingual{Off}{关} & \bilingual{Path}{路径} & \bilingual{Boundary supported}{有边界} & \Second{\MechRolePathBoundaryGPA{}} & \Second{\MechRolePathBoundaryExact{}} & \Second{\MechRolePathBoundaryNLL{}} \\
\rowcolor{OurRowGray}\bilingual{On}{开} & \bilingual{Path}{路径} & \bilingual{Endpoint only}{仅端点} & \Best{\MechCompleteEndpointGPA{}} & \Best{\MechCompleteEndpointExact{}} & \Best{\MechCompleteEndpointNLL{}} \\
\rowcolor{OurRowGray}\bilingual{On}{开} & \bilingual{Path}{路径} & \bilingual{Boundary supported}{有边界} & \Best{\MechCompleteBoundaryGPA{}} & \Best{\MechCompleteBoundaryExact{}} & \Best{\MechCompleteBoundaryNLL{}} \\
}
\newcommand{\MechanismStrataErrorRows}{%
\bilingual{Off}{关} & \bilingual{Mean}{期望} & \bilingual{Endpoint only}{仅端点} & \MechRoleMeanEndpointDup{} & \MechRoleMeanEndpointMiss{} & \MechRoleMeanEndpointDelay{} \\
\bilingual{Off}{关} & \bilingual{Mean}{期望} & \bilingual{Boundary supported}{有边界} & \MechRoleMeanBoundaryDup{} & \MechRoleMeanBoundaryMiss{} & \MechRoleMeanBoundaryDelay{} \\
\bilingual{On}{开} & \bilingual{Mean}{期望} & \bilingual{Endpoint only}{仅端点} & \Best{\MechPhaseMeanEndpointDup{}} & \MechPhaseMeanEndpointMiss{} & \Second{\MechPhaseMeanEndpointDelay{}} \\
\bilingual{On}{开} & \bilingual{Mean}{期望} & \bilingual{Boundary supported}{有边界} & \Second{\MechPhaseMeanBoundaryDup{}} & \MechPhaseMeanBoundaryMiss{} & \Second{\MechPhaseMeanBoundaryDelay{}} \\
\bilingual{Off}{关} & \bilingual{Path}{路径} & \bilingual{Endpoint only}{仅端点} & \MechRolePathEndpointDup{} & \Second{\MechRolePathEndpointMiss{}} & \MechRolePathEndpointDelay{} \\
\bilingual{Off}{关} & \bilingual{Path}{路径} & \bilingual{Boundary supported}{有边界} & \MechRolePathBoundaryDup{} & \Second{\MechRolePathBoundaryMiss{}} & \MechRolePathBoundaryDelay{} \\
\rowcolor{OurRowGray}\bilingual{On}{开} & \bilingual{Path}{路径} & \bilingual{Endpoint only}{仅端点} & \Second{\MechCompleteEndpointDup{}} & \Best{\MechCompleteEndpointMiss{}} & \Best{\MechCompleteEndpointDelay{}} \\
\rowcolor{OurRowGray}\bilingual{On}{开} & \bilingual{Path}{路径} & \bilingual{Boundary supported}{有边界} & \Best{\MechCompleteBoundaryDup{}} & \Best{\MechCompleteBoundaryMiss{}} & \Best{\MechCompleteBoundaryDelay{}} \\
}

%% file: shared/tables/closure_rows.tex
\newcommand{\ObjectRows}{%
$\lambda_i=0$ & \bilingual{Count only}{仅计数} & \Second{\ObjectCountOnlyEndpointSoftUMAE{}} & \Second{\ObjectCountOnlyEndpointHardUMAE{}} & \Second{\ObjectCountOnlyEndpointSoftGainMAE{}} & \Second{\ObjectCountOnlyEndpointHardGainMAE{}} \\
$\lambda_i=0$ & \bilingual{Identity verified}{身份核验} & \Second{\ObjectCountOnlyVerifiedSoftUMAE{}} & \Second{\ObjectCountOnlyVerifiedHardUMAE{}} & \Second{\ObjectCountOnlyVerifiedSoftGainMAE{}} & \Second{\ObjectCountOnlyVerifiedHardGainMAE{}} \\
\rowcolor{OurRowGray}\bilingual{Complete}{完整} & \bilingual{Count only}{仅计数} & \Best{\ObjectIdentityEndpointSoftUMAE{}} & \Best{\ObjectIdentityEndpointHardUMAE{}} & \Best{\ObjectIdentityEndpointSoftGainMAE{}} & \Best{\ObjectIdentityEndpointHardGainMAE{}} \\
\rowcolor{OurRowGray}\bilingual{Complete}{完整} & \bilingual{Identity verified}{身份核验} & \Best{\ObjectIdentityVerifiedSoftUMAE{}} & \Best{\ObjectIdentityVerifiedHardUMAE{}} & \Best{\ObjectIdentityVerifiedSoftGainMAE{}} & \Best{\ObjectIdentityVerifiedHardGainMAE{}} \\
}
\newcommand{\ObjectIdentityRows}{%
$\lambda_i=0$ & \Second{\ObjectCountOnlyVerifiedNewPrecision{}} & \Second{\ObjectCountOnlyVerifiedNewRecall{}} & \Second{\ObjectCountOnlyVerifiedReentryDouble{}} & \Second{\ObjectCountOnlyVerifiedUnresolved{}} & \Second{\ObjectCountOnlyVerifiedDelay{}} \\
\rowcolor{OurRowGray}\bilingual{Complete}{完整} & \Best{\ObjectIdentityVerifiedNewPrecision{}} & \Best{\ObjectIdentityVerifiedNewRecall{}} & \Best{\ObjectIdentityVerifiedReentryDouble{}} & \Best{\ObjectIdentityVerifiedUnresolved{}} & \Best{\ObjectIdentityVerifiedDelay{}} \\
}
\newcommand{\MatchedAccessRows}{%
\rowcolor{OurRowGray}Full & \Best{\MatchedTimestampFullGPA{}} & \Best{\MatchedTimestampFullExact{}} \\
\rowcolor{OurRowGray}Stream & \Second{\MatchedTimestampStreamGPA{}} & \Second{\MatchedTimestampStreamExact{}} \\
}
\newcommand{\SelectionRows}{%
\bilingual{Source train pool}{来源训练池} & \PoolTrain{} \\
\bilingual{OVO media exclusion}{OVO媒体排除} & \ExcludedOVOQueries{} \\
\bilingual{Dependency-family exclusion}{依赖族排除} & \ExcludedDependencyQueries{} \\
\bilingual{Graph-incompatible exclusion}{图不可达排除} & \ExcludedGraphQueries{} \\
\bilingual{Other unsupported records}{其他无支持记录} & \ExcludedUnsupportedQueries{} \\
\bilingual{Selected training queries}{最终训练查询} & \SelectedQueries{} \\
}
\newcommand{\SupervisionRows}{%
\bilingual{Endpoint only}{仅端点} & \SelectedCountOnly{} \\
\bilingual{Verified identity}{身份核验} & \SelectedIdentity{} \\
\bilingual{Verified boundary}{边界核验} & \SelectedBoundary{} \\
\bilingual{Verified stable interval}{稳定区间核验} & \SelectedStable{} \\
}
\newcommand{\VisualAuditRows}{%
\bilingual{Objects}{物体} & \AuditObjectsN{} & \AuditObjectsEndpoint{} & \AuditObjectsIncrement{} \\
\bilingual{Natural events}{自然事件} & \AuditNaturalN{} & \AuditNaturalEndpoint{} & \AuditNaturalIncrement{} \\
\bilingual{Periodic}{周期} & \AuditPeriodicN{} & \AuditPeriodicEndpoint{} & \AuditPeriodicIncrement{} \\
}
\newcommand{\ReachabilityRows}{%
Action & \ReachActionQueries{} & \ReachActionFull{} & \ReachActionStream{} \\
Transit & \ReachTransitQueries{} & \ReachTransitFull{} & \ReachTransitStream{} \\
Episode & \ReachEpisodeQueries{} & \ReachEpisodeFull{} & \ReachEpisodeStream{} \\
Periodic & \ReachPeriodicQueries{} & \ReachPeriodicFull{} & \ReachPeriodicStream{} \\
}

%% file: sections/00_abstract.tex
Continuous video counting requires distinguishing new observations from new objects or completed events. We introduce \StaMina{} (\StaMinaName{}), which learns to maintain counting state through state-conditioned updates. Recurrent visual context supports recognition; learned transitions maintain visibility, persistent identities, and completed-event records. A differentiable recurrence trains event transitions over legal paths constrained by count endpoints; visibility and association objectives train the object branch. A multi-source pipeline organizes \SpatialQueriesK{}K spatial queries and complementary event annotations into counting trajectories. On SVCBench, we evaluate counting adaptation with partial video overlap and held-out groups of linked annotations. Under prefix replay (Full) and persistent streaming (Stream), 4B and 8B models reach \RStaFourFullOverall{}/\RStaFourStreamOverall{} and \RStaEightFullOverall{}/\RStaEightStreamOverall{} Gaussian Precision Accuracy, respectively. The 8B model gains \MatchedEightGainFull{}/\MatchedEightGainStream{} points over Counting-SFT on the same queries. Matched-graph comparisons isolate phase conditioning and trajectory supervision, assessing training objectives alongside hard decisions. Online video benchmarks and count-conditioned decisions assess online understanding and task eligibility.

%% file: sections/en_body.tex
\section{Introduction}
A packing assistant sees a blue striped mug and an orange mug beside a box (Figure~\ref{fig:overview}a). A mug that disappears from view and reappears is still the same object: visibility changes while the number of distinct mugs stays fixed. The orange mug is then lowered into the box. The placement remains in progress while the hand holds it and completes after release. Observing the placed mug again preserves that completed-event count. These cases distinguish changes in visibility, identity, and completion \citep{svcbench,yang2025thinking,shangguan2025tomato}.

Long-context and streaming vision-language models retain visual evidence through attention, recurrent summaries, or learned associative memory \citep{bai2025qwen3,xu2025streamingvlm,flashvstream,livestar,ttt}. Hybrid local attention and online learners extend this context along a stream \citep{streamttt}. The inventory clerk in Figure~\ref{fig:overview}b illustrates the additional role of a record: remembered appearance supports recognition, while the notebook indicates which placements have already been counted. Thus the same observation can warrant an update during an active event and a hold after completion (Figure~\ref{fig:overview}c). The key is to learn when current evidence changes previously recorded state.

We introduce \StaMina{} (\StaMinaName{}), which learns to maintain the counting state through state-conditioned transitions. Gated delta memory carries causal appearance and motion context; typed state records current visible candidates, accepted identities, and completed events. An event head uses visual context and the current phase to score legal updates. Training sums the probabilities of paths consistent with supported count endpoints, allowing interval counts to supervise completion and persistence without assigning a boundary to every frame. The object branch separately learns visibility and association, so that an identity decision can be deferred while a visible mug remains countable.

This formulation connects data construction to the decisions being learned. The training resource must cover both identity-preserving reappearance and the transition from held to placed. We construct such supervision from visual tracking, independent prefix counting, and verified periodic transformations. The pool contains \SpatialQueries{} spatial queries and complementary event records. Count endpoints constrain the number of accepted updates; reliable identities and boundaries supervise their local meaning. Snapshot and interval queries share the underlying trajectory.

We instantiate \StaMinaFour{} and \StaMinaEight{} using the corresponding Qwen3-VL~\citep{bai2025qwen3} instruction models. Our contributions are:
\begin{itemize}\setlength{\itemsep}{1pt}\setlength{\parskip}{0pt}
\item \textbf{State-conditioned causal counting.} A recurrent visual context and typed transition heads distinguish identity birth, visibility change, and event completion, with numerical answers read from the accepted state.
\item \textbf{Trajectory-constrained transition learning.} A differentiable forward recurrence trains event transitions from supported count endpoints, while verified boundaries and identities supervise their respective operators.
\item \textbf{Trajectory data and evaluation.} A multi-source pipeline supplies \SpatialQueriesK{}K spatial queries and complementary event trajectories. Evaluation at two model scales covers counting adaptation, online-video transfer on disjoint media, and count-conditioned action predicates.
\end{itemize}

\section{Related Work}
\paragraph{Streaming visual context.}
Long-video benchmarks test temporal evidence, retrieval, and spatial reasoning \citep{fu2025video,wu2024longvideobench,svcbench,niu2025ovo,li2026ovosbench,liu2026trace,kwan2026video,fan2026medclaw}. Online evaluation also ties answers to the observations available at a decision time \citep{lin2024streamingbench,niu2025ovo}. Streaming architectures maintain context through memory consolidation, decoupled perception, or selective retention \citep{qian2024streaming,qian2025dispider,xu2025streamingvlm,flashvstream,livestar,zeng2025streamforest}. We study the subsequent decision that converts this context into a persistent count update.

\paragraph{Associative memory and state transitions.}
Test-time training (TTT) adapts parameters online \citep{sun2020test,gandelsman2022test,videottt,xie2025test,nath2026clipttt}; fast-weight layers use these parameters as recurrent state \citep{ttt,tandon2025end,dalal2025one}. StreamTTT uses nonlinear fast weights with short-window attention \citep{streamttt}; Spatial TTT adds geometry-aware prediction \citep{spatialttt}; Cambrian-S studies persistent spatial recall and counting \citep{cambrians}. Our continuous branch uses gated linear delta updates \citep{gateddelta,kimilinear}. The counting mechanism resides in the state-conditioned transition model and its trajectory objective. ActionSwitch learns action-state changes with a conservative transition loss \citep{actionswitch}. Connectionist temporal classification (CTC) marginalizes compatible alignment paths \citep{graves2006ctc}, while neural video alignment learns from weak action annotations \citep{nnviterbi}. We constrain operator-specific paths by multiple count endpoints, carry phase posteriors between them, and evaluate causal integer commitments.

\paragraph{Counting data and embodied state.}
VSI-Bench, TOMATO, CountLLM, and AV-Reasoner study spatial reasoning, temporal reasoning, and repetitive or clue-grounded counting \citep{yang2025thinking,shangguan2025tomato,yao2025countllm,lu2025av}. SVCBench distinguishes eight counting operators and their trajectories \citep{svcbench}. We construct aligned supervision from ScanNet, ScanNet++, ARKitScenes, and Ego4D \citep{dai2017scannet,yeshwanth2023scannetpp,baruch2021arkitscenes,grauman2022ego4d}. Count-conditioned predicates connect the resulting state to action eligibility, complementing observation-to-control models \citep{openvla}.

\section{StaMina: Learning to Maintain Counting State}
\label{sec:method}

\begin{figure}[t]
\centering
\includegraphics[width=\linewidth]{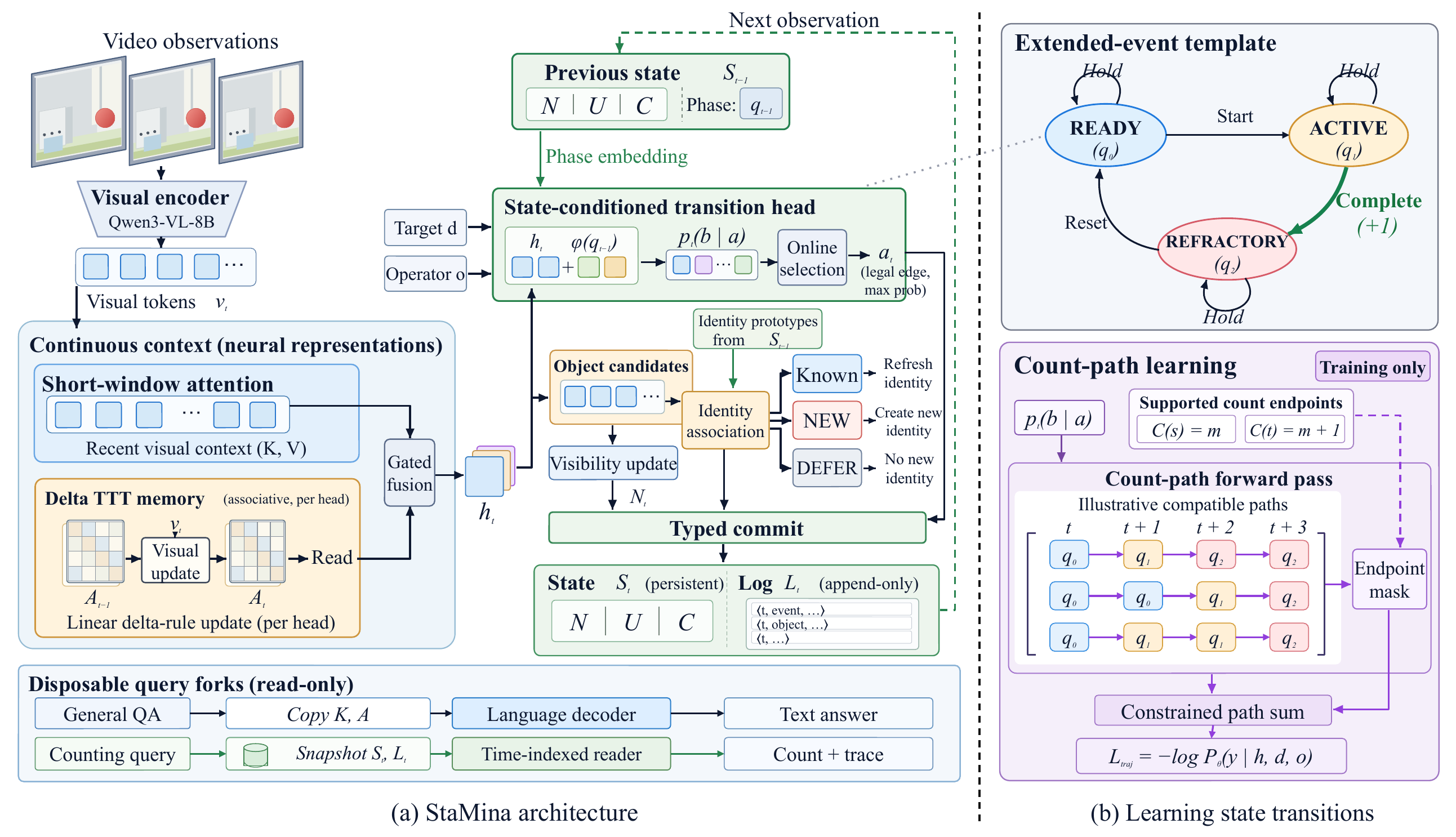}
\vspace{-10pt}
\caption{\textbf{StaMina architecture and state-transition learning.} (a) Short-window attention and delta memory supply causal features for event transitions and independent visibility/identity updates. DEFER postpones identity association while preserving visible candidates. Accepted updates maintain counting state and an evidence log, with separate readouts for general QA and counting. Previous state conditions transition decisions. (b) The event graph defines legal changes. Training sums the probabilities of paths matching supported count endpoints; online inference selects a legal outgoing edge at each observation.}
\label{fig:fsm}
\vspace{-5pt}
\end{figure}

Figure~\ref{fig:fsm} summarizes the causal context, typed state updates, and count-trajectory learning in StaMina.

\subsection{Counting semantics and causal context}
For target description $d$, let $N_d(t)$, $U_d(t)$, and $C_{d,o}(t)$ denote the true visible, first-seen, and completed-event counts, respectively. For $s<t$, O1-Snap/Delta ask for $N_d(t)$/$N_d(t)-N_d(s)$; O2-Unique/Gain ask for $U_d(t)$/$U_d(t)-U_d(s)$. Event operators specify which completions count. Hats denote model predictions. Observations have timestamps $\tau_k$; a query at $t$ reads step $k(t)=\max\{k:\tau_k\le t\}$. The target vector $e_d$ is the normalized mean of its pretrained token embeddings.

The runtime maintains recent key--value (KV) context $K$, associative memory $A$, typed state $S$, and an accepted-transition log $L$. Counting targets are registered at the trajectory origin; later targets require prefix replay. General QA uses the language decoder, whereas counting reads $S$. Query generation uses disposable forks and leaves the visual stream unchanged.

For visual-token input $x_i$, each memory head projects $k_i=W_kx_i$, $v_i=W_vx_i$, and $q_i=W_qx_i$ into the same width. Hats on keys and queries denote unit normalization. The gated delta update is
\begin{align}
 \bar A_i&=\gamma_i A_{i-1},&
 A_i&=\bar A_i+\eta_i(u_i-\bar A_i\hat k_i)\hat k_i^\top,\label{eq:ttt}\\
 r_i&=\hat q_i+\operatorname{LN}(A_i\hat q_i),&
 u_i&=\operatorname{LN}(v_i-k_i),\nonumber
\end{align}
where $\operatorname{LN}$ is layer normalization \citep{ba2016layernorm}; sigmoid retention $\gamma_i$ and write $\eta_i$ gates lie in $(0,1)$. Slow parameters learn projections and gates; $A_i$ stores stream-specific associations. Gated fusion combines this linear online learner \citep{gateddelta,kimilinear} with short-window attention to produce causal observation features $h_k$. Memory continues updating during repeated observations and unfinished actions.

\subsection{State-conditioned event transitions}
\label{sec:fsm}
For operator $o$, a finite-state machine (FSM) defines phases $\mathcal Q_o$ and legal edges $\mathcal E_o$. A nonlinear head scores visual context, target, operator embedding $e_o$, phases, and edge role $r(a,b)$ (hold, start, complete, or reset):
\begin{equation}
 p_k^o(b\mid a)=\operatorname{softmax}_{b:(a,b)\in\mathcal E_o}
 g_\theta(h_k,e_d,e_o,e_a,e_b,e_{r(a,b)}).
 \label{eq:transition}
\end{equation}
For an extended event, READY permits a start, ACTIVE permits completion, and REFRACTORY requires reset before another event. Every phase permits hold; only completion increments the count. Other operators use the templates in Appendix~\ref{app:training}. Causally associated candidates retain separate phases; concurrent counts use their joint graph.

All candidate phases share the same $h_k$ during training. Deployment selects the highest-scoring outgoing edge, including hold. Phase feedback enters the score rather than the continuous-memory writer: the same completion evidence can support a change during an event and a hold after its acceptance.

\subsection{Identity and visibility transitions}
Let $\mathcal V_k$ contain deduplicated visible candidates and $\mathcal I_k$ accepted persistent identities. Target-conditioned pooling predicts visibility and association to a retrieved identity, NEW, or DEFER. A visible DEFER candidate remains in $\mathcal V_k$ without creating an identity. One-to-one assignment matches existing identities; NEW inserts an identity. Disappearance removes visibility while retaining identity. Predictions are
\begin{equation}
 \widehat N_d(t)=|\mathcal V_{k(t)}|,\qquad
 \widehat U_d(t)=|\mathcal I_{k(t)}|.
 \label{eq:objects}
\end{equation}
Thus, a reappearing mug can be visible before its identity is resolved. A delayed NEW decision increments $\widehat U$ at acceptance, so first-seen timing errors can affect both Unique and Gain. Soft visibility and birth totals receive count supervision; verified identities supervise association. Training rolls the same predicted hard registry as deployment, with discrete assignments detached. Appendix~\ref{app:training} specifies these losses.

\subsection{Learning from count trajectories}
\label{sec:learning}
Let $\delta_o(a,b)\ge0$ be an edge's count increment. The probability of legal paths ending at phase $b$ and count $n$ is
\begin{equation}
 F_k(b,n)=\sum_{a:(a,b)\in\mathcal E_o}
 F_{k-1}(a,n-\delta_o(a,b))p_k^o(b\mid a).
 \label{eq:forward}
\end{equation}
At each supported endpoint $(k_j,y_j)$, retain the slice $n=y_j$, sum over phases, and normalize before continuing. The product of normalization factors gives the joint probability of all supported endpoints:
\begin{equation}
 \mathcal L_{\mathrm{traj}}=-\log P_\theta(\mathbf y\mid\mathbf h,d,o).
 \label{eq:trajectory}
\end{equation}
An interval increment of two constrains paths to two completions without assigning unverified frame boundaries. Equal cumulative endpoints constrain completion-free paths; equal visible counts can still contain entries and exits.

The objective combines answer-token cross-entropy, event trajectories, soft object counts, verified identity matches, and verified phase-edge cross-entropy:
\begin{equation}
 \mathcal L=\mathcal L_{\mathrm{ans}}+\lambda_p\mathcal L_{\mathrm{traj}}+
 \lambda_o\mathcal L_{\mathrm{object}}+\lambda_i\mathcal L_{\mathrm{id}}+
 \lambda_b\mathcal L_{\mathrm{boundary}}.
 \label{eq:loss}
\end{equation}
Each term uses its available supervision sites; positive weights and masks are specified in Appendix~\ref{app:training}. Phase persists across supervised intervals. Path likelihood rewards mass on correct-count paths, whereas deployment selects one edge per observation. Their counts and completion times can differ; Section~\ref{sec:dataexp} pairs endpoint fit with hard-prefix correctness and delay, including count-only and verified-supervision strata.

\subsection{Online commitment and readout}
An accepted update commits to $(S,L)$ with an evidence ID and acceptance timestamp \citep{eventsource}. Repeated delivery of that ID is idempotent; association and phase inference determine whether distinct observations depict the same instance. Observation and effective timestamps both equal $\tau_k$, and queries read the latest committed state at or before their endpoint. Interval answers subtract two versions of the same trajectory. Acceptance never rewrites an earlier answer.

KV and associative matrices have fixed capacity; identity indexes and accepted-event logs grow outside that context. For $T$ observations and count support width $B$, event training costs $O(T|\mathcal E_o|B)$, using the joint graph for concurrent slots. Online decoding scores outgoing edges from the committed phase. Sampling times, rather than token chunks, determine which events are observable.

\section{Trajectory-Based Counting Data}
\label{sec:data}
The data pipeline provides the update supervision required by Section~\ref{sec:learning}. Each trajectory fixes a source segment, target, operator, and count origin; its parent video determines the provenance and split membership. A supervision family links annotations within a parent through shared entities or events, temporal transformations, and count-derived answers. Segment-local counts reset at that segment's origin, while queries within the segment share state.

\subsection{Video-to-trajectory construction}
A proposal stage discovers countable objects and event descriptions from raw videos. For spatial targets, segmentation and tracking provide candidate visual evidence, while independent VLM prefix views estimate visible and cumulative counts. The acceptance stage checks agreement, supported changes, and conflicting re-identifications. Accepted records store $N_d(t)$ or $U_d(t)$ with their evidence references. These are quality-filtered automatic labels; stable instance IDs are retained where supported, and a fused count is not treated as an identity annotation.

Natural-event records use causal prefix counting and support checks for action, transition, and episode targets. Periodic records apply authorized temporal transformations to annotated source cycles, retaining the frame map and transformed completion boundaries. This gives count-trajectory supervision for all operators and boundary supervision for the subset with verified timing.

\begin{figure}[t]
\centering
\includegraphics[width=\linewidth]{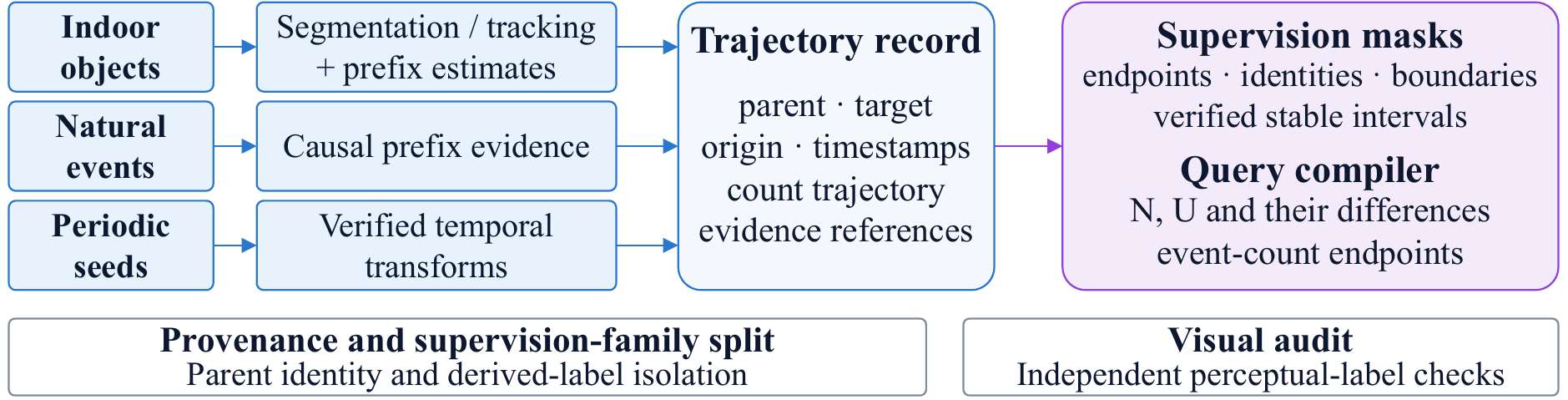}
\caption{\textbf{Counting-data construction.} Source-specific evidence checks produce canonical trajectories and causal queries, with inherited provenance and supervision-family exclusion.}
\label{fig:data}
\end{figure}

\subsection{Causal query compilation and auditing}
The query compiler samples supported endpoints and emits the target description, query time, causal media scope, and integer answer. O1-Delta is derived from two supported $N_d$ endpoints; O2-Gain is derived from two supported $U_d$ endpoints. Verified stable intervals are retained to supervise persistence; equal visible-count endpoints alone do not establish a stable interval. Rejected or unsupported observations are omitted rather than converted into zero counts.

An independent checker recomputes the differences, validates temporal order, checks duplicate IDs and media hashes, and verifies the parent key of every derived record. This establishes algebraic and provenance consistency. Perceptual label quality is assessed separately through held-out audits and the training ablations in Section~\ref{sec:training-data}.

\subsection{Dataset composition and splits}
The multi-source counting pool combines indoor spatial videos, natural action videos, and transformed periodic sequences. The pool contains \PoolGroups{} grouped records, \PoolQueries{} query--answer pairs, \PoolParents{} parent groups, and \PoolMedia{} referenced media files. The spatial portion contains \SpatialQueries{} queries; natural events contribute \NaturalQueries{} and periodic transformations contribute \PeriodicQueries{}. The training and development splits of this pool contain \PoolTrain{} and \PoolDev{} queries. Appendix~\ref{app:data} gives the per-operator inventory.

Training and development splits group records by source video and scene. SVCBench~\citep{svcbench} evaluates counting adaptation with partial source-video overlap. Held-out supervision families exclude their counts, derived intervals, and linked identity/boundary evidence across operators and temporal transformations (Appendix~\ref{app:data}). Both online-video evaluation sets and their derived media are jointly excluded from training and model selection. The inventory describes the source pool; Appendix~\ref{app:closure-results} gives the selected training membership.

\input{sections/05_experiments}

%% file: sections/05_experiments.tex
\section{Experiments}
\label{sec:experiments}
\subsection{Experimental setup}
\paragraph{Models and evaluation.}
We train \StaMinaFour{} and \StaMinaEight{} from Qwen3-VL-4B/8B-Instruct \citep{bai2025qwen3}. Counting-SFT denotes answer-token supervised fine-tuning on the same selected counting queries, without the added memory or state heads. Controls share visual-token and update budgets within each comparison. SVCBench covers \OperatorCount{} operators, \BenchVideos{} videos, \BenchQuestions{} questions, and \BenchQueries{} timestamped queries \citep{svcbench}, using the supervision-family split in Section~\ref{sec:data} for counting adaptation. OVO-Bench and OVO-S-Bench evaluate online understanding and spatial reasoning through the language decoder \citep{niu2025ovo,li2026ovosbench}. One checkpoint per model size and seed is used across all three benchmarks. Appendix~\ref{app:training} fixes the model and optimization settings; Appendix~\ref{app:data} specifies training membership and supervision-family exclusion.

\paragraph{Access and scoring.}
For our models, Full resets the runtime per query and replays its causal prefix using unwindowed attention over at most \FullFrameBudget{} frames, or \OVOSFullFrames{} on OVO-S. Stream processes new observations at \StreamFPS{} fps, retaining a \StreamKVBudget{}-token window, fast weights, and typed state. Both modes run the state interface. Counting targets are registered at the trajectory origin; matched controls receive them at the same time. General QA follows each benchmark's question-reveal schedule. Gaussian Precision Accuracy (GPA) averages query scores within each question, then across questions. Monotonicity Consistency (MoC) and Update Detection Accuracy (UDA) evaluate cumulative trends and responses at true changes. Our benchmark scores average the unrounded scores of \TestSeeds{} training seeds. Every official query remains in the score, including graph-unreachable endpoints. Appendix~\ref{app:reference} gives metric and reference-protocol details. Throughout, gray identifies the complete method; bold and underline mark the best and second-best values within the indicated comparison groups.

\subsection{Counting adaptation}
The complete 8B method gains \MatchedEightGainFull{}/\MatchedEightGainStream{} GPA points over Counting-SFT in Full/Stream on shared counting queries (Table~\ref{tab:paired}). This measures the joint contribution of state structure, trajectory objectives, and auxiliary supervision; Section~\ref{sec:dataexp} isolates transition learning. The 4B/8B models in Table~\ref{tab:main} reach \RStaFourFullOverall{}/\RStaEightFullOverall{} in Full and \RStaFourStreamOverall{}/\RStaEightStreamOverall{} in Stream. Both Full scores exceed the highest proprietary reference, Gemini-3-Flash (\RGeminiOverall{}).
\MainComparison

Scaling adds \FourToEightFullGain{}/\FourToEightStreamGain{} points. Full exceeds Stream by \FourAccessGap{} and \EightAccessGap{} points at the two sizes. These gaps compare complete access settings, including their different observation schedules. A matched-timestamp replay/persistence comparison in Appendix~\ref{app:closure-results} holds observations and target availability fixed.

\subsection{Learning state transitions}
\label{sec:dataexp}
\paragraph{Identifying the learned decision.}
The event-head comparison caches causal features, candidate associations, and lifecycle masks from a common training-only warmup snapshot. With these inputs frozen, it crosses phase input with expected-count or path-likelihood training and shares verified-boundary supervision. A separate annotated diagnostic set supports exact transition errors and paired comparisons; its construction is given in Appendix~\ref{app:closure-results}. Table~\ref{tab:mechanism} shows that phase conditioning and the path objective both improve hard prefix counts. Their combination also reduces false updates on verified stable intervals. Hard exact match rises from \MechRolePathExact{} to \MechCompleteExact{} when phase conditioning is added to path training. The result concerns which legal transition is selected, since legal eligibility is unchanged across all rows.
\begin{table}[tb]
\centering\small
\caption{\textbf{Event-head controls on frozen 8B Stream features.} The diagnostic inputs and legal graph are fixed. GPA, hard prefix exact match, and false-update rate are percentages; a lower false-update rate is better.}
\label{tab:mechanism}
\setlength{\tabcolsep}{6pt}
\begin{tabular}{llccc}
\toprule Phase input & Endpoint objective & GPA $\uparrow$ & Hard exact $\uparrow$ & False updates $\downarrow$\\\midrule
\MechanismRows
\bottomrule
\end{tabular}
\end{table}

\paragraph{From endpoint fit to accepted counts.}
Each control produces endpoint likelihood and hard prefix accuracy from the same checkpoint and causal rollout. The improvements persist on trajectories from endpoint-only sources, while verified boundaries provide a stronger timing signal. Phase conditioning can improve hard counts even when its mean-objective negative log-likelihood (NLL) changes little; endpoint fit and online accuracy therefore carry complementary information. Duplicate completions, missed increments, and acceptance delay jointly test whether fewer updates reflect more accurate decisions. Appendix~\ref{app:closure-results} reports both the supervision strata and the paired intervals. The system-level no-FSM comparison, which also changes the legal graph, is reported separately in Figure~\ref{fig:ablations}.

\begin{figure}[tb]
\centering
\includegraphics[width=\linewidth]{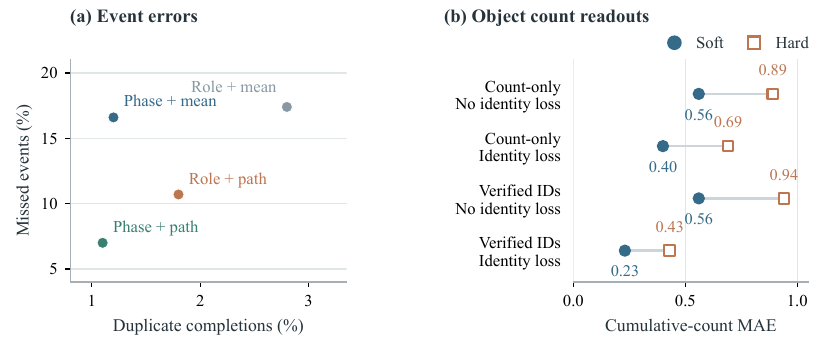}
\caption{\textbf{Hard-decision diagnostics at 8B.} Event errors distinguish repeated or missed completions from stable-interval updates. Object endpoints compare the differentiable cumulative count with the accepted identity count under the same rollout. Detailed denominators and supervision strata are given in Appendix~\ref{app:closure-results}.}
\label{fig:commitment}
\end{figure}

\paragraph{Visibility and identity.}
Object diagnostics compare count supervision with and without the verified identity term, retaining predicted registry rollout in both cases. Soft and hard count errors are evaluated on the count-only and identity-verified source strata. Figure~\ref{fig:commitment} connects the soft/hard object gap to event errors. Identity supervision reduces hard cumulative-count error alongside re-entry double-counting and unresolved identities. Cross-query Gain errors include identities accepted after their first appearance; final totals alone do not assess this timing error. Identity precision and delay use verified tracks, while count-only records remain in the endpoint-error evaluation.

\subsection{Training data analysis}
\label{sec:training-data}
\paragraph{Supervision quality and coverage.}
The selected training set contains \SelectedQueries{} queries from \SelectedParents{} parents, including \SelectedIdentity{} identity-supported and \SelectedBoundary{} boundary-supported queries. Visual endpoint correctness ranges from \AuditNaturalEndpoint{}\% on natural events to \AuditPeriodicEndpoint{}\% on periodic records. Appendix~\ref{app:closure-results} reports the selected training queries, overlapping source videos, available supervision masks, and stratified visual audits. Endpoint accuracy and interval accuracy are assessed separately from algebraic consistency. Training exclusions are reported alongside an observation-budget capacity check on official evaluation endpoints. All official scores retain endpoints above this capacity, exposing the effect of sparse observations on dense cycles.

\paragraph{Why trajectories matter.}
At matched token and update budgets, complete trajectories improve over general video QA by \FactoryGainFull{}/\FactoryGainStream{} points. Single-query training removes cross-query persistence supervision. The temporal control permutes visual observations over fixed timestamp slots while retaining endpoints, supervision masks, and candidate associations. This training corruption breaks visual--update alignment while preserving graph feasibility; evaluation uses the original sequence. Consistency filtering tests label selection. Table~\ref{tab:data} reports these comparisons. Together with the supervision audit, they connect the data contribution to learning updates rather than to the number of QA rows alone.

\subsection{Online understanding and efficiency}
\label{sec:transfer}
\paragraph{Capabilities after counting adaptation.}
The language decoder answers the original OVO-Bench and OVO-S questions. Table~\ref{tab:transfer} summarizes current perception, historical evidence, response timing, and spatial reasoning. The 8B model reaches \OVOEightFullAvg{}/\OVOEightStreamAvg{} on OVO-Bench and \OSSEightFullAvg{}/\OSSEightStreamAvg{} on OVO-S. Gains concentrate on historical tracing and spatiotemporal context tracking; real-time perception decreases slightly, while spatial reasoning changes are mixed. All comparisons share the checkpoint and training selection for each model and seed. StreamForest and Flash-VStream retain their native protocols \citep{zeng2025streamforest,flashvstream2024}; within-family initial-model controls and full task breakdowns appear in Appendix~\ref{app:transfer}.
\TransferComparison

\paragraph{Cost and count-conditioned decisions.}
For general QA, mean query latency is \LatencyStream{} s in Stream and \LatencyFull{} s in Full. Appendix~\ref{app:complete-runtime} distinguishes query generation, continuous ingestion, deterministic count readout, and replay for a newly revealed counting target. The fixed GPU context is complemented by a host identity index and event log. On recorded inventory, handover, restocking, and repetition videos, fixed count predicates translate state errors into eligibility decisions. The two 8B conditions share the predicates and balanced trials; exact results and decision definitions appear in Table~\ref{tab:embodied} and Appendix~\ref{app:interventions}.

\section{Conclusion}
\StaMina{} learns to maintain the counting state through state-conditioned transitions and trajectory supervision. Matched-graph and soft/hard diagnostics link these decisions to accepted counts. A multi-source pipeline supplies endpoint, identity, and timing supervision, while the shared language decoder supports online video understanding.

%% file: sections/90_ai_use.tex
\section*{AI Use Statement}
We used generative AI tools to assist with language editing and improve the clarity of the manuscript, as well as to identify relevant literature. Vision-language models also assisted automatic annotation in the data-construction pipeline described in Section~\ref{sec:data}. The authors take responsibility for the accuracy, originality, and integrity of the final manuscript.

\section*{Reproducibility Statement}
Section~\ref{sec:method} specifies the state representation, causal memory, and learned count-state transitions of \StaMina{}. Appendix~\ref{app:training} details the architecture, training objectives, hyperparameters, and runtime procedure. Section~\ref{sec:data} and Appendix~\ref{app:data} describe data construction and supervision-family exclusion. Appendices~\ref{app:reference}, \ref{app:mechanism-controls}, and~\ref{app:complete-runtime} document the evaluation protocol, comparison settings, and computational accounting. Together, these descriptions support implementation and comparison under the stated settings. We will release the code and model checkpoints to the research community.

\section*{Ethics Statement}
This work studies causal video counting and count-conditioned action predicates using existing datasets, benchmarks, and pretrained models. Use of these resources should respect their licenses and access conditions. Because errors in visual evidence and accumulated counts can affect downstream decisions, applications should include task-specific safety evaluation, appropriate operational safeguards, and human oversight suited to the deployment context.

%% file: sections/99_appendix.tex
\section{Training and Runtime Specification}
\label{app:training}
\StaMinaFour{} and \StaMinaEight{} initialize from Qwen3-VL-4B/8B-Instruct. Their processor, visual tokenization, and pretrained widths are unchanged. A delta branch parallels attention in every \CfgMemoryStride{}th decoder block, before the feed-forward sublayer. Each branch concatenates \CfgMemoryHeads{} heads of width \CfgMemoryWidth{}, projects their readouts to the decoder width, and adds a sigmoid-gated residual to the attention output. Each square $A_0$ is a shared trainable parameter initialized to zero and copied at a new trajectory. Retention and write gates are sigmoids of separate linear projections of the token input, scalar per head. The channel-wise retention comparison changes this gate alone \citep{kimilinear}. Slow weights are fixed at deployment. Scoring heads use the Gaussian error linear unit (GELU) activation \citep{hendrycks2016gelu}.

\begin{table}[h]
\centering\small
\caption{Default complete-system configuration for both model scales. Pretrained dimensions follow Qwen3-VL; the frozen-input event-head controls use the training protocol in Appendix~\ref{app:mechanism-controls}.}
\label{tab:model-config}
\begin{tabular}{lc}
\toprule Setting & Value\\\midrule
Memory heads $\times$ width; block stride & $\CfgMemoryHeads{}\times\CfgMemoryWidth{}$; \CfgMemoryStride{}\\
Transition/association hidden width & \CfgHeadWidth{} (GELU)\\
Pooling queries; entity-event slots & \CfgPoolQueries{}; \CfgEventSlots{}\\
Identity shortlist; visibility threshold & \CfgIdentityTopK{}; \CfgVisibilityThreshold{}\\
Deduplication/track cosine thresholds & \CfgDedupCosine{}/\CfgAssociationCosine{}\\
Token chunk; continuous truncation & \CfgChunkTokens{}; \TruncationK{} chunks\\
Warmup; joint epochs; trajectory batch & \WarmupSteps{}; \CfgJointEpochs{}; \CfgTrajectoryBatch{}\\
Decoder/projector; new-module learning rates & \texttt{\CfgBaseLR{}}; \texttt{\CfgHeadLR{}}\\
AdamW $(\beta_1,\beta_2)$; $\epsilon$ & $(\CfgAdamBetaOne{},\CfgAdamBetaTwo{})$; \texttt{\CfgAdamEpsilon{}}\\
Weight decay; gradient-norm clip & \CfgWeightDecay{}; \CfgGradClip{}\\
$(\lambda_p,\lambda_o,\lambda_i,\lambda_b)$ & $(\CfgLambdaPath{},\CfgLambdaObject{},\CfgLambdaIdentity{},\CfgLambdaBoundary{})$\\\bottomrule
\end{tabular}
\end{table}

\paragraph{Features and nonlinear scores.}
The target vector $e_d$ is the unit-normalized mean of target-token input embeddings; $e_o$, phase embeddings, and edge-role embeddings are learned. Final decoder features of the current observation provide a mean visual vector and target-conditioned candidate vectors. For an associated candidate, their concatenation is $h_k$; a global target uses the mean vector in both positions. Equation~\ref{eq:transition} is instantiated as
\begin{align}
 z_k&=W_hh_k+W_de_d+W_oe_o,\\
 s_k(a,b)&=w^\top\operatorname{GELU}(z_k+W_ae_a+W_be_b+W_re_{r(a,b)}+c).
\end{align}
Operator, phase, and role embeddings and all score projections have width \CfgHeadWidth{}. A legal-edge softmax produces $p_k^o(b\mid a)$. Zeroing $e_a,e_b$ retains edge roles and the legal mask. Projection matrices in different branches are independent even when their subscripts coincide. Features, target encoding, and candidate association are independent of the event path being summed.

\paragraph{Candidate lifecycle and event graphs.}
E2-Periodic always uses a single global cycle stream. For the other event operators, the target schema fixes global or entity-associated counting before observations arrive: global actions use one stream and entity-associated actions use at most \CfgEventSlots{} slots. Learned target-conditioned pooling queries attend to current visual features. Candidates are ordered by visibility, with query index breaking ties; cosine duplicate suppression retains distinct feature vectors. Candidate--slot association maximizes summed cosine similarity subject to one-to-one matches above \CfgAssociationCosine{}. Equal-score assignments use candidate order and then slot-creation order. An unmatched visible candidate takes the first unused slot; excess candidates are recorded as overflow. Slot prototypes are replaced by detached matched features. Slots retain their phase and prototype through absence and are not recycled within a segment. An absent slot has only hold enabled. Association and creation are causal, detached, and identical across mechanism controls.

\begingroup
\raggedright
New slots start READY. E1-Action completes on $\mathrm{READY}\to\mathrm{REFRACTORY}$ and re-arms on observed release/renewed eligibility. E1-Transit uses the same graph: reaching the requested destination completes; re-establishing its source resets. E2-Episode uses $\mathrm{READY}\to\mathrm{ACTIVE}\to\mathrm{REFRACTORY}\to\mathrm{READY}$, with completion only on $\mathrm{ACTIVE}\to\mathrm{REFRACTORY}$. E2-Periodic uses $\mathrm{READY}\to\mathrm{AWAY}$ for departure and $\mathrm{AWAY}\to\mathrm{READY}$ for the specified return, which completes one cycle. Every phase allows hold, and there is no extra time guard. Initial phases never use annotation labels; incomplete observation of a left-censored event is evaluated under the same initialization.\par
\endgroup

For concurrent slots $\ell$, the joint transition is
\begin{equation}
 p_k(\mathbf b\mid\mathbf a)=\prod_\ell p_{k\ell}(b_\ell\mid a_\ell),\qquad
 \delta(\mathbf a,\mathbf b)=\sum_\ell\delta_o(a_\ell,b_\ell).
\end{equation}
The forward state retains the joint phase tuple after conditioning on a shared count, so posterior dependencies between slots are preserved. Candidate association is fixed within this sum. Hard decoding maximizes this product from the committed tuple, equivalently selecting each slot's best outgoing edge. Ties prefer hold, then the template edge order. Same-time completions across slots are summed. With $m$ allocated slots, these templates have at most $3^m$ phase tuples and $2^m$ outgoing edges per tuple.

\paragraph{Sampled time and reachable counts.}
Let source-frame times be $\rho_j$. If a prefix contains $M$ frames and its budget is $B_f$, Full takes $m=\min(M,B_f)$ frames at indices $\lfloor r(M-1)/(m-1)\rfloor$, $r=0,\ldots,m-1$; for $m=1$, it takes index zero. Stream takes the latest source frame at each grid time from the segment origin at \StreamFPS{} fps, suppressing duplicate frame indices. Observation timestamps $\tau_k$ are the selected source-frame times. Query and interval boundaries introduce no extra frames. A read at $s$ uses $k(s)$, or the empty initial state if no observation precedes $s$. Full interval endpoints are read from the same replay used for that query. Labels retain their physical query times: a change between the last sample and a query remains a scored error if unobserved. Accepted changes have observation and effective time $\tau_k$; later evidence cannot backdate an answer.

Each slot traverses at most one edge per observation. The global periodic stream therefore completes at most $\lfloor T/2\rfloor$ cycles in $T$ observations. A single E1 slot completes at most $\lceil T/2\rceil$ events, and an episode slot at most $\lfloor(T+1)/3\rfloor$. Allocating all \CfgEventSlots{} entity slots from the first observation gives the corresponding aggregate upper ceilings. The run checks reachable support using actual timestamps, slot creation/absence, and initialization. Graph-incompatible training endpoints are logged and excluded. Every official evaluation query remains scored, including overflow and unreachable endpoints. Increasing the count array cannot recover unobserved transitions.

\paragraph{Endpoint and boundary supervision.}
Initially $F_0(a,n)=\mathbf1[a=a_{\mathrm{READY}},n=0]$ for the joint ready state; inactive slots are constrained to hold until creation. Endpoint $j$ supplies $Z_j=\sum_aF_{k_j}(a,y_j)$. We retain $F_{k_j}(a,y_j)/Z_j$ and minimize $-\sum_j\log Z_j$, carrying its phase posterior onward. After conditioning, subtracting $y_j$ from the count coordinate preserves its offset and phase posterior. Count support spans all reachable interval increments and computations use log probabilities. For an interval-only label, the count coordinate starts at zero at the interval's left endpoint while inheriting the marginal phase distribution from the preceding observations. Interval-only chains are chronological and non-overlapping; overlapping alternatives use separate causal replays and share their supervision-family weight. Interval labels directly supervise count increments. Shared endpoints and derived differences form one supervision family rather than repeated independent losses.

Verified phase-edge records form $\mathcal M_B=\{(k,\ell,a^*,b^*)\}$. A physical transition at $\tau^*$ maps to the first sampled observation with $\tau_k\ge\tau^*$. A site is used only if the associated instance and both phases are verified and $(\tau_{k-1},\tau_k]$ contains exactly that legal edge, or verified persistence for a hold. Multiple hidden transitions, uncertain phases, and unmatched instances mask out the site. The loss is conditional edge cross-entropy,
\begin{equation}
 \mathcal L_{\mathrm{boundary}}=-\frac{1}{|\mathcal M_B|}
 \sum_{(k,\ell,a^*,b^*)\in\mathcal M_B}\log p_{k\ell}^o(b^*\mid a^*).
\end{equation}
It uses causal features at $\tau_k$; boundary times and phases are supervision, not runtime inputs. No observation after a query is inserted to satisfy a boundary label. Verified hold and change sites receive equal sampling weight. An empty mask contributes no loss. Count-only records use endpoint slices. The recurrence is exact for the given causal features, association, and event graph; endpoint fit and greedy commitment are evaluated separately.

\paragraph{Visibility, association, and delayed identity acceptance.}
The object branch uses \CfgPoolQueries{} target-conditioned pooling queries. With current visual vectors $v_{kj}$ and learned queries $q_i$, its candidate is $x_{ki}=\sum_j\alpha_{kij}v_{kj}$, where $\alpha_{kij}\propto\exp[(q_i+W_de_d)^\top W_vv_{kj}/\sqrt{D}]$ and $D=\CfgHeadWidth{}$ is the attention width. Visibility is $\nu_{ki}=\sigma(w_v^\top x_{ki}+b_v)$. A candidate is suppressed when its cosine similarity to an earlier survivor is at least \CfgDedupCosine{}, before the visibility threshold \CfgVisibilityThreshold{}. Hard visibility uses the surviving candidates above that threshold; soft losses sum over all deduplicated candidates before thresholding. Entity-event pooling uses the same construction.

For each candidate, the \CfgIdentityTopK{} nearest historical prototypes by cosine similarity form $\mathcal R_{ki}$ (all available identities if fewer). The association multilayer perceptron (MLP) has the same hidden width as the transition head. The operator $\operatorname{sg}$ preserves its input value and stops its gradient:
\begin{equation}
\begin{aligned}
 p_{ki}(c)&=\operatorname{softmax}_{c\in\mathcal R_{ki}\cup\{\mathrm{NEW},\mathrm{DEFER}\}}
 g_{\mathrm{id}}(x_{ki},e_d,\bar q_c),\\
 \bar q_c&=\begin{cases}
 \operatorname{sg}(q_c), & c\in\mathcal R_{ki},\\
 q_c, & c\in\{\mathrm{NEW},\mathrm{DEFER}\}.
 \end{cases}
\end{aligned}
\end{equation}
The score is $g_{\mathrm{id}}=w_{\mathrm{id}}^\top\operatorname{GELU}(W_xx+W_de_d+W_q\bar q_c+c_{\mathrm{id}})$, with independent projections. NEW and DEFER have trainable class embeddings with the same width as candidate features and stored prototypes. For visible candidates, one-to-one assignment maximizes the sum of log probabilities, with a private NEW/DEFER alternative per candidate. Each existing identity receives at most one match. Equal-score assignments are resolved in candidate order, preferring DEFER, then the oldest identity, then NEW. Matches replace prototypes with detached features; NEW appends an identity; DEFER keeps visibility without insertion. Persistent identities are never evicted within a trajectory. A deferred candidate is reconsidered from current evidence when observed again.

A visible unresolved object immediately contributes to $\widehat N$. An old-identity resolution leaves $\widehat U$ unchanged; NEW increments it at acceptance. The target $U$ counts physical first appearances, so delayed acceptance can yield $\widehat U(t)-\widehat U(s)\ne U(t)-U(s)$ even after the final unique count becomes correct. We measure both acceptance delay and cross-query Gain errors without rewriting earlier prefixes.

\paragraph{Object losses and gradient boundaries.}
Soft visible and birth totals are
\begin{equation}
 \widetilde N_k=\sum_i\nu_{ki},\quad
 \widetilde b_k=\sum_i\nu_{ki}p_{ki}(\mathrm{NEW}),\quad
 \widetilde U_k=\sum_{j\le k}\widetilde b_j.
 \label{eq:softobjects}
\end{equation}
The birth accumulator resets at the count origin. It is a surrogate conditioned on predicted identity history, not a marginalization of identity paths. Let $\mathcal J_N,\mathcal J_U$ index supported visible and unique endpoints and $\mathcal M_{\mathrm{id}}$ verified candidate/identity labels. Then
\begin{align}
 \mathcal L_{\mathrm{object}}&=\operatorname{mean}_{j\in\mathcal J_N}(\widetilde N_{k_j}-y_j^N)^2+\operatorname{mean}_{j\in\mathcal J_U}(\widetilde U_{k_j}-y_j^U)^2,\\
 \mathcal L_{\mathrm{id}}&=-\operatorname{mean}_{(k,i)\in\mathcal M_{\mathrm{id}}}\log p_{ki}(c^*_{ki}).
\end{align}
Empty sets contribute no term; interval labels use corresponding soft differences. An identity label is admitted only for a unique verified match inside the shortlist or a verified first appearance eligible for NEW. Ambiguity, missing identities, and retrieval misses mask out identity CE and never become NEW or DEFER labels. Retrieval misses are recorded separately in the error log. $\mathcal L_{\mathrm{ans}}$ is the mean answer-token cross-entropy on supported causal QA records. Counting answers also supervise this decoder during training; deployed counting still reads typed state.

Training advances the same predicted hard registry as deployment. Selection, deduplication, retrieval, assignment, insertion, and historical writes are detached. Gradients reach retained current features and visibility/association heads, including the NEW and DEFER embeddings; soft accumulators remain connected until supervised endpoints. No count label substitutes for a predicted identity. Soft/hard count error is measured in both count-only and identity-verified source strata. NEW precision/recall, re-entry double counting, unresolved rate, and acceptance delay use verified identities; unresolved cases remain failures rather than disappearing from the denominator.

\paragraph{Optimization and resets.}
The vision encoder stays frozen. During \WarmupSteps{} warmup updates, only new memory branches and pooling/transition/association heads are trained with available answer, object, identity, and boundary terms. Joint training follows the Stream observation and KV schedule and updates those modules, the decoder, and visual projector for \CfgJointEpochs{} epochs with the full objective, cosine learning-rate decay to zero, and the AdamW \citep{loshchilov2019adamw} settings in Table~\ref{tab:model-config}. A batch contains \CfgTrajectoryBatch{} trajectories; token, object, identity, and boundary losses average over available sites within each trajectory, while the trajectory term retains its joint NLL. Each term is then averaged over contributing trajectories. Each complete-system final checkpoint is used across all benchmarks. Appendix~\ref{app:mechanism-controls} defines the separate frozen-input event-head controls.

Continuous-memory backpropagation spans \TruncationK{} token chunks while forward state persists. The event forward graph and soft object accumulators remain connected to their supervised trajectory losses; optimizer updates follow the whole trajectory batch. At a visual step, context updates, scores are computed, and hard changes commit before query readout. Query forks are discarded. A new segment/count origin resets KV, slots, identities, logs, and matrices to their initial states; queries inside it do not reset state. Full reconstructs the prefix with unwindowed causal attention; Stream persists bounded context and typed state.

\section{Data Inventory and Quality Semantics}
\label{app:data}
\begin{table}[h]
\centering\small
\caption{Composition of the multi-source pool by counting operator. Entries count query--answer pairs, not independent source videos.}
\label{tab:inventory}
\begin{tabular}{lc}
\toprule Operator & Queries\\\midrule
O1-Snap & \PoolSnap{}\\
O1-Delta & \PoolDelta{}\\
O2-Unique & \PoolUnique{}\\
O2-Gain & \PoolGain{}\\
E1-Action & \PoolAction{}\\
E1-Transit & \PoolTransit{}\\
E2-Episode & \PoolEpisode{}\\
E2-Periodic & \PoolPeriodic{}\\\midrule
Total & \PoolQueries{}\\\bottomrule
\end{tabular}
\end{table}
A grouped record contains a question and a sequence of supported query times and counts. Expanding it into SFT rows changes the physical row count but not the number of underlying trajectories. Delta and Gain reuse endpoints from their source trajectory. Multiple media files can be crops or temporal transformations of one parent. We therefore retain parent groups, grouped records, query points, and referenced media as separate inventory units.

Spatial and natural-event labels are automatically supported count trajectories. They provide supervision for count changes and persistence. Periodic transformations retain verified frame maps from annotated seeds and support temporal boundary training. Algebraic replay checks derived answers, while visual audits assess whether the source evidence supports the counts. Unsupported targets and failed generation attempts remain outside the accepted training pool.

\paragraph{Supervision acceptance and selection.}
Every accepted endpoint links to its causal media scope and source evidence. Finite integer counts, ordered local timestamps, consistent endpoint differences, and operator-valid trends are required. Conflicting visual evidence disables the affected supervision mask; algebraic agreement alone does not create identity or boundary labels. Periodic boundaries additionally require the transformed frame map. Visual audits separately assess exact endpoint and interval-increment accuracy, stratified by source and operator.

A trajectory key contains parent, derived segment/frame map, canonical target, operator, and count origin. Exclusion operates on supervision families before SFT expansion. For one parent, aliases and overlapping entity/event scopes are linked across every timestamp, crop, speed change, and repeated-cycle derivative. The dependency graph links $N$ to Delta, $U$ to Gain, shared identity tracks to both object families, and event counts to their supporting phase/boundary annotations. Overlapping operator definitions and any QA answer derived from these nodes join the same connected component. If a component intersects a held-out benchmark target, all its count, interval, identity, boundary, and answer-token supervision is excluded from training and model selection. Consequently, hiding Gain alone cannot leave its two Unique endpoints available. Uncertain semantic matches are excluded rather than separated by wording.

SVCBench evaluates adaptation to held-out supervision families with partial source-video overlap. The family exclusions apply to all project-specific training stages and model selection, including warmup, answer fine-tuning, and transition training. Training and development records are grouped by source scene and video before these exclusions. For transfer, the union of OVO-Bench and OVO-S evaluation media, original source videos, and all derived media is removed from every training and development selection. One final checkpoint per scale and seed is then shared across all three benchmarks. Each run records family IDs, ancestry, exclusions, selected queries, and supervision-mask counts; the source-pool inventory is reported separately.

\paragraph{Target-reveal protocol.}
For counting, the target description and operator are registered at the trajectory origin; future answers and completion boundaries are not runtime inputs. Matched controls receive the same target information. Each target/operator trajectory is evaluated separately, so a resident-target timing measurement does not imply an unbounded number of concurrent targets. A late target replays the available prefix to construct its state, and that replay cost is included in its end-to-end latency. General QA follows each benchmark's original question-reveal schedule.

\section{Reference Models and Scoring}
\label{app:reference}
Table~\ref{tab:main} includes the complete reference roster from SVCBench's camera-ready object and event tables \citep{svcbench}. Human performance and the text-only GPT-4-Turbo control \citep{openai2023devday} provide context. The proprietary group contains Gemini-3-Flash \citep{google_gemini3_flash2025}, Doubao-Seed-1.8 \citep{doubao_seed18}, and GPT-5.4 \citep{openai_gpt54_model_2026}. The open-weight group contains Kimi-K2.5 \citep{team2026kimi}, Qwen3-VL-8B/30B and Qwen2.5-VL-7B \citep{bai2025qwen3,qwen25vl}, InternVL-3.5-8B \citep{wang2025internvl3}, Molmo2-8B \citep{clark2026molmo2}, Qwen3.5-35B-A3B \citep{qwen3_5}, the streaming models StreamingVLM, Dispider, LiveStar, and Flash-VStream-7B \citep{xu2025streamingvlm,qian2025dispider,livestar,flashvstream}, and StaMina. Model groups reflect weight availability; the Access column specifies Full or Stream. Within each access mode, reference models and StaMina adaptation rows are ranked separately across weight-availability groups.

Qwen3.5-35B-A3B retains its published aggregate over \ReferenceQwenScoredQuestions{} scored questions, including \ReferenceQwenPeriodicQuestions{} periodic questions rather than the nominal \ReferenceNominalPeriodicQuestions{}. Human performance also retains its published aggregate, rather than a reconstruction from rounded categories.

Full denotes per-query access to a causal video prefix; Stream denotes sequential ingestion with persistent state. Each model retains its input sampling protocol. Most Full references use the benchmark's \FullFrameBudget{}-frame sampling protocol, while Gemini-3-Flash uses native \StreamFPS{} fps input. The Stream references follow their native \StreamFPS{} fps inference. Reference scores come from the original SVCBench evaluations; Table~\ref{tab:paired} compares system variants on shared counting queries.

\begin{equation}
 \mathrm{GPA}_q=\frac{1}{n_q}\sum_i\exp\!\left[-\frac{(\hat y_{qi}-y_{qi})^2}{2\sigma_{qi}^2}\right],
 \qquad \sigma_{qi}=0.05\max(y_{qi},1).
\end{equation}
For GPA, first average query-point scores within each question, then average across questions. Overall is consequently weighted by category question counts, not an equal mean of eight category cells. MoC applies to O2-Unique and the four cumulative event categories. UDA applies to O1-Snap, O2-Unique, and the four event categories. Single-query differences do not have trajectory metrics. UDA evaluates changes on ground-truth change steps; it does not measure false updates during unchanged steps. Stable-interval false updates are evaluated in the event diagnostics.

Our SVCBench results use the arithmetic mean of unrounded per-seed question scores over \TestSeeds{} seeds; all own SVCBench conditions score the common \BenchQuestions{} questions, with category weights determined by their question counts. Means and differences are computed before display rounding. The confidence intervals shown for the two event-head contrasts use \BootstrapReplicates{} parent-level resamples and paired predictions. Parent/query IDs, invalid flags, actual frame times, checkpoint identities, and scoring code identify each comparison. Published reference aggregates retain their source evaluation populations.

\section{Intervention and Embodied Evaluation Details}
\label{app:interventions}
The no-FSM condition is retrained with an independent increment head and commits above the fixed confidence threshold \CfgVisibilityThreshold{}. It removes phase conditioning and the legal-transition constraint while retaining the same continuous memory, supported intervals, and verified-boundary records. Count-state readout controls use the complete model's fixed rollout. Clear replaces its count projections and readout log with the empty state. Delay maps each requested endpoint to the previous distinct read endpoint for the same target, including interval starts, or to the empty origin when none exists. Both interval endpoints use that same mapping. Swap fixes a different-video donor with a compatible object or event schema, preferring an identical target when available. Stream reads the donor's persistent state at the recipient's elapsed times; Full uses the latest donor prefix no later than the recipient query, or the empty state if unavailable. Both interval endpoints use one donor history. These probes test the indexed count payload, including its availability; visual processing and neural state updates remain fixed.

Each routine uses \PredicateParents{} independent recorded videos with \PredicateQueries{} decisions per video, giving \PredicateTrials{} decisions balanced between eligible and ineligible cases. These parents are disjoint from training, all three benchmarks, and the core diagnostic collections. The fixed predicates are inventory $N\ge\PredicateInventoryThreshold{}$, handover $C\ge\PredicateHandoverThreshold{}$, restocking $N\le\PredicateRestockThreshold{}$, and repetition completion $C\ge\PredicateCycleThreshold{}$. Verified video counts determine eligibility at the fixed thresholds above. Counting-SFT and the complete 8B model both use Stream access and answer at the same timestamps; their integer counts enter the same predicate function. Invalid or unavailable counts are scored as incorrect. For each model, accuracy pools one prediction per decision within each routine, dividing the number of correct eligibility decisions by \PredicateTrials{}.

\section{Transition-Learning Comparisons}
\label{app:mechanism-controls}
Table~\ref{tab:mechanism} crosses phase conditioning with the endpoint objective on fixed causal inputs; detailed results appear in Appendix~\ref{app:closure-results}. For each 8B seed, the common training-only warmup snapshot is frozen before the four heads diverge. Its chronological Stream pass caches observation and target features, candidate associations, slot creation/absence masks, and thus the actual legal graph and count support. Training and diagnostic caches use the same frozen snapshot; diagnostic labels never enter this pass. The recurrent memory still updates causally while creating the cache.

All heads start from the same warmup scorer parameters. Only the transition scorer and its operator, phase, and role embeddings are optimized; the decoder, memory, pooling, and association parameters remain frozen. The four conditions use the same ordered batches of \CfgTrajectoryBatch{} training trajectories and the same cached observation/token inputs, endpoint labels, boundary masks, initial phases, and greedy decoder. Each receives \CfgMechanismSteps{} updates at initial learning rate \texttt{\CfgHeadLR{}}, with fresh AdamW states and cosine decay to zero; other optimizer settings follow Table~\ref{tab:model-config}. Phase-off zeroes $e_a,e_b$ and retains semantic edge roles and legal eligibility. The expected-count control uses unconditioned forward marginals $\bar F$, without endpoint-label slicing:
\begin{equation}
 \mu_j=\sum_{a,n}n\bar F_{k_j}(a,n),\qquad
 \mathcal L_{\mathrm{mean}}=\operatorname{mean}_{j\in\mathcal J}(\mu_j-y_j)^2.
\end{equation}
For interval labels, the expected increment is the difference of the two unconditioned means. Head optimization uses $\lambda_p\mathcal L_{\mathrm{mean}}+\lambda_b\mathcal L_{\mathrm{boundary}}$ or $\lambda_p\mathcal L_{\mathrm{traj}}+\lambda_b\mathcal L_{\mathrm{boundary}}$, with the same weights; answer, object, and identity terms have no trainable parameters in this control. The phase-conditioned path head instantiates the complete event-transition rule under fixed inputs. Its control checkpoint is distinct from the jointly trained benchmark model, so this comparison identifies transition learning conditional on the shared visual evidence. Independent increments, fixed phase filters, and memory-cell comparisons remain separate complete-system controls.

\paragraph{Diagnostic populations and commitment.}
Diagnostic trajectories are independent of the training and model-selection parents. Endpoint-only and boundary-supervised strata describe the source's available training annotations; all diagnostic endpoints are verified for scoring and are never optimization inputs. Within each event-head condition, the same control checkpoint and cached causal inputs produce endpoint NLL and greedy hard-prefix predictions. Reported NLL divides the joint conditional negative log-likelihood by the number of verified endpoints. Predictions are matched chronologically within a slot to the nearest unmatched true completion in the preceding \DiagnosticEventMatchWindow{} seconds. Additional predictions within a true-completion window are duplicates; unmatched true completions are misses and predictions outside all windows are false updates. Stable false-update rate counts verified sampling intervals containing any commit, excluding intervals within \DiagnosticEventMatchWindow{} seconds on either side of a true completion. False-update rate divides marked stable intervals by all eligible stable intervals; duplicate and miss rates divide their counts by ground-truth events. These ratios pool counts across trajectories and seeds. Delay averages prediction time minus matched completion time over all matched events; unmatched events remain misses. Parent-level paired intervals quantify the factorial contrasts. Low NLL and poor hard counts identify a remaining surrogate/decoder gap; delay is reported alongside accuracy so conservative holds cannot appear successful merely by suppressing updates.

Object controls follow complete-system joint training, setting $\lambda_i$ to zero or retaining the complete objective. Registry-update rules are shared, and each condition rolls its own predicted identity history. Both count-only-source and identity-verified-source records contribute soft/hard Unique error and supported Gain error. NEW precision/recall, re-entry duplication, unresolved identities, and acceptance delay use the verified subset, and retrieval misses are identified in the error log. Soft/hard Unique MAE uses supported endpoints; Gain MAE uses consecutive endpoint differences, including the empty pre-observation origin before the first query. NEW precision counts the first accepted birth of a true identity as a true positive and repeated or unmatched births as false positives; recall divides true positives by the number of true identities. Re-entry duplication is measured per re-entry encounter. Unresolved rate sums seen-but-unaccepted identities over queries and divides by all seen-identity exposures at those queries. Identity ratios pool their numerators and denominators across trajectories and seeds. Delay averages acceptance minus first-seen time over all resolved identities, with unresolved cases retained in the unresolved rate. Cross-query errors retain the original first-seen target time even when identity is accepted later. Soft birth fitting and hard identity acceptance are compared empirically rather than equated.

\paragraph{Observations and temporal controls.}
The matched-access experiment uses the jointly trained benchmark checkpoints. Replay and persistence receive the same causal frames, target revelation, and queries. It measures the access implementation, including replay versus persistence and unwindowed versus bounded attention, after removing the standard sampling difference. The capacity diagnostic compares each ground-truth endpoint with an observation-budget ceiling: $\lfloor T/2\rfloor$ for the global periodic stream, $\CfgEventSlots{}\lceil T/2\rceil$ for E1, and $\CfgEventSlots{}\lfloor(T+1)/3\rfloor$ for episodes. Full uses its frame-budget ceiling; Stream has at most $T=\lfloor(t-t_0)f\rfloor+1$ observations for origin $t_0$ and sampling rate $f$. Coverage below these ceilings is an upper bound on feasible endpoint coverage; actual slot availability, discarded duplicate frames, and joint endpoint constraints can reduce it. Speed/pause transformations preserve event identity and map boundaries through their frame ancestry. With the target and observation schedule fixed, Stream state updates are independent of when that state is queried. Standard Full resamples and replays each queried prefix, so changing query times can change its visual processing. Whole-action duplication increases the annotated count. Prefix decoding, target replay, and reset rules follow Appendix~\ref{app:training} in every condition.

\section{Scope and Failure Modes}
Candidate overflow, missed observations, and identity association errors can precede transition scoring. Fixed event slots preserve phase through absence but bound the represented entity streams. The accepted-identity index and event log grow with distinct objects and accepted events, even while neural context stays bounded. Endpoint likelihood and soft births are training objectives; online integer correctness and acceptance timing are measured on hard rollouts. The transfer benchmarks test the shared checkpoint on excluded media, while SVCBench measures the declared same-video new-question adaptation.

%% file: sections/95_closure.tex
\section{Mechanism and Data Diagnostics}
\label{app:closure-results}
\paragraph{Diagnostic collections.}
The event collection contains \DiagEventTrajectories{} trajectories from \DiagEventParents{} parent videos, balanced across the four event operators. The object collection contains \DiagObjectTrajectories{} trajectories from \DiagObjectParents{} parents. Each trajectory has \DiagQueriesPerTrajectory{} endpoint queries. These additional videos come from the source collections used by the pipeline and are excluded from training and development before model fitting; they are not counted in the source-pool inventory. Every condition uses the same parents, observations, targets, query times, and \TestSeeds{} training seeds.

Event strata identify whether the source pipeline supplies count endpoints alone or also verified boundaries. Dense temporal annotations are added to both diagnostic strata for evaluation and are never used to fit their models. Object count-only sources have verified visible and cumulative endpoints; identity-supported sources additionally have verified physical tracks. Identity metrics use the latter group, while both groups contribute count errors. Each collection is balanced between its two source strata. This grouping distinguishes available training supervision from evaluation annotations.

\paragraph{Paired reporting.}
GPA, hard exact match, endpoint-normalized NLL, and mean absolute error (MAE) average within trajectories and then across trajectories and seeds. Ratio metrics instead pool numerator and denominator across trajectory--seed records: false updates use eligible stable intervals; duplicates and misses use ground-truth events; identity ratios use their specified eligible identity units. Delay pools all matched completions or resolved identities. Parent-level resampling retains all trajectories and all seed predictions belonging to a selected parent. The two event-head contrasts use the same resampled parent indices for both conditions and \BootstrapReplicates{} replicates. Their intervals describe variation across parents conditional on the evaluated seed set. The factorial event study trains separate head-only control checkpoints on common frozen inputs. The complete object and matched-access evaluations use the end-to-end benchmark checkpoint; the identity-loss ablation is trained separately. With path likelihood fixed, phase conditioning adds \MechPhaseEffectGain{} GPA points (\DiagConfidencePercent{}\% interval: \MechPhaseEffectLow{}--\MechPhaseEffectHigh{}); with phase conditioning fixed, path likelihood adds \MechPathEffectGain{} points over expected-count training (\MechPathEffectLow{}--\MechPathEffectHigh{}). These are conditional contrasts on the common frozen inputs.

\paragraph{Event decisions and endpoint fit.}
Table~\ref{tab:hardevents} complements Table~\ref{tab:mechanism} with endpoint negative log-likelihood (NLL), duplicate completions, missed events, and acceptance delay. NLL is computed with the same edge probabilities that drive greedy decoding. Event matching is performed chronologically within an instance; a matched completion contributes a delay, an unmatched target remains a miss, and excess accepted completions remain errors. Verified intervals with no event provide the false-update denominator. Delay is reported together with misses, so postponing every update cannot produce an apparently good result. Table~\ref{tab:event-strata} separates endpoint-only and boundary-supported sources.

\begin{table}[tb]
\centering\small
\caption{Event hard decisions and trajectory fit. GPA, exact match, duplicates, and misses are percentages. NLL is normalized by the number of supervised endpoints and delay is in seconds.}
\label{tab:hardevents}
\setlength{\tabcolsep}{4pt}
\begin{tabular}{llcccccc}
\toprule Phase & Objective & GPA $\uparrow$ & Exact $\uparrow$ & NLL $\downarrow$ & Duplicate $\downarrow$ & Miss $\downarrow$ & Delay $\downarrow$\\\midrule
\MechanismDetailRows
\bottomrule
\end{tabular}
\end{table}

\begin{table}[tb]
\centering\small
\caption{Event diagnostics by source supervision. Evaluation annotations are held out; each source stratum is ranked separately.}
\label{tab:event-strata}
\setlength{\tabcolsep}{4pt}
\begin{tabular}{lllccc}
\toprule Phase & Objective & Source stratum & GPA $\uparrow$ & Exact $\uparrow$ & NLL $\downarrow$\\\midrule
\MechanismStrataRows
\bottomrule
\end{tabular}
\par\smallskip
\begin{tabular}{lllccc}
\toprule Phase & Objective & Source stratum & Duplicate $\downarrow$ & Miss $\downarrow$ & Delay $\downarrow$\\\midrule
\MechanismStrataErrorRows
\bottomrule
\end{tabular}
\end{table}

\paragraph{Object counts and identity acceptance.}
The object ablation sets $\lambda_i=0$ and keeps all other training terms, masks, budgets, and registry operations fixed. Table~\ref{tab:object-hard} reports mean absolute errors for soft and hard cumulative counts and their interval differences. Table~\ref{tab:identity-hard} evaluates NEW decisions and re-entry behavior on verified identities. An unresolved identity stays in the recall and unresolved-rate denominators. Gain is evaluated between adjacent endpoints, including the empty pre-observation origin before the first query, so a late birth can incur an interval error even when the final unique count matches. Both conditions use the same candidate budget and selection rules.

\begin{table}[tb]
\centering\small
\caption{Soft and hard object errors. Each column reports MAE ($\downarrow$) on the same endpoint or interval queries; ranks compare the two training conditions within each source stratum.}
\label{tab:object-hard}
\begin{tabular}{llcccc}
\toprule Identity loss & Source stratum & Soft $U$ & Hard $U$ & Soft Gain & Hard Gain\\\midrule
\ObjectRows
\bottomrule
\end{tabular}
\end{table}

\begin{table}[tb]
\centering\small
\caption{Identity decisions on verified tracks. Precision, recall, re-entry double-counting and unresolved rates are percentages; delay is in seconds.}
\label{tab:identity-hard}
\setlength{\tabcolsep}{4pt}
\begin{tabular}{lccccc}
\toprule Identity loss & NEW precision $\uparrow$ & NEW recall $\uparrow$ & Re-entry $\downarrow$ & Unresolved $\downarrow$ & Delay $\downarrow$\\\midrule
\ObjectIdentityRows
\bottomrule
\end{tabular}
\end{table}

\paragraph{Matched observation schedules.}
Table~\ref{tab:matched-access} supplies identical timestamps and target availability to Full replay and Stream persistence on the event diagnostic set. Full uses unwindowed prefix attention while Stream retains its fixed window and recurrent context. This comparison measures the remaining access differences with sampling controlled; it is distinct from the official-budget results in Table~\ref{tab:main}.
\begin{table}[tb]
\centering\small
\caption{Matched-timestamp access on the event diagnostic collection. Both conditions use the complete 8B model.}
\label{tab:matched-access}
\begin{tabular}{lcc}
\toprule Access & GPA $\uparrow$ & Hard exact $\uparrow$\\\midrule
\MatchedAccessRows
\bottomrule
\end{tabular}
\end{table}

\paragraph{Training selection and supervision coverage.}
Table~\ref{tab:training-selection} separates the source pool from the selected training membership. The selected data span \SelectedParents{} parents; \SelectedSVCParents{} overlap SVCBench and contribute \SelectedSVCQueries{} retained queries from new supervision families. Selection removes evaluation-media descendants, held-out supervision families and graph-incompatible training endpoints in that order, recording each removal once. Parent overlap with SVCBench describes visual exposure; family-level exclusion removes linked counts, intervals, identities and boundaries. A Gain interval cannot be held out while its determining Unique endpoints remain in training. Counts linked by overlapping intervals or the same physical identity/event evidence belong to one exclusion component, including transformed time coordinates.

\begin{table}[tb]
\centering\small
\caption{Training selection from the source pool. Counts refer to queries unless a row specifies parents.}
\label{tab:training-selection}
\begin{tabular}{lc}
\toprule Selection statistic & Count\\\midrule
\SelectionRows
\bottomrule
\end{tabular}
\end{table}

All selected queries carry supported count labels. Table~\ref{tab:supervision-coverage} partitions them into endpoint-only, identity-supported object, and boundary-supported event records. Verified stability overlaps these strata and is reported separately, so its count is not added to the training-set size. Auxiliary labels require the corresponding verified evidence.
\begin{table}[tb]
\centering\small
\caption{Supervision coverage in the selected training set. The first three strata are mutually exclusive; verified stability overlaps them.}
\label{tab:supervision-coverage}
\begin{tabular}{lc}
\toprule Supervision & Selected queries\\\midrule
\SupervisionRows
\bottomrule
\end{tabular}
\end{table}

\paragraph{Visual quality and reachability.}
The audit samples grouped records by source group, restricted to records with both supported endpoints and positive-duration intervals. Each sampled record contributes one endpoint and one interval, drawn independently from its eligible sets without reference to answer correctness; zero-count increments remain eligible. Two annotators inspect causal clips and reconcile disagreements before assigning the binary correctness label. Identity and boundary masks additionally require the physical track or transition time to be supported. Table~\ref{tab:visual-audit} reports sample counts and correctness rates; it evaluates perception rather than algebraic self-consistency. Table~\ref{tab:reachability} checks the necessary count-capacity bound before prediction, using the maximum observation budget and ignoring candidate failures. Full permits at most \FullFrameBudget{} observations; Stream permits at most $\lfloor t\,\StreamFPS{}\rfloor+1$ from the zero origin. A global periodic stream completes at most $\lfloor T/2\rfloor$ cycles. Other event bounds allow all \CfgEventSlots{} entity slots to be present from the origin. Passing this bound does not establish joint-path or perceptual reachability. Every official query remains in the main score.
\begin{table}[tb]
\centering\small
\caption{Stratified visual-label audit. Each record contributes one endpoint and one interval judgment; both rates use the audited-record count as denominator.}
\label{tab:visual-audit}
\begin{tabular}{lccc}
\toprule Source & Audited records & Endpoint correct (\%) & Increment correct (\%)\\\midrule
\VisualAuditRows
\bottomrule
\end{tabular}
\end{table}
\begin{table}[tb]
\centering\small
\caption{Observation-budget capacity coverage on official event endpoints (\%). This is an upper bound on reachable coverage; all endpoints remain in the official score.}
\label{tab:reachability}
\begin{tabular}{lccc}
\toprule Operator & Endpoint queries & Full upper coverage & Stream upper coverage\\\midrule
\ReachabilityRows
\bottomrule
\end{tabular}
\end{table}

\section{Complete-System Runtime}
\label{app:complete-runtime}
The runtime evaluation uses the complete 8B model on the same NVIDIA H200 host, BF16 arithmetic, and batch size \RuntimeBatch{}. General-QA latency averages requests generating \RuntimeOutputTokens{} output tokens. Stream begins with the visual prefix already processed and takes \LatencyStream{} seconds per request; Full includes its \FullFrameBudget{}-frame causal prefix replay and takes \LatencyFull{} seconds. These are mean request latencies, with no seed-level confidence interval attached.

Chronological visual ingestion runs at \StreamIngestFPS{} frames per second, including the vision encoder, causal memory, and typed-state update for a registered counting target. A typed-count read takes \CountReadMs{} milliseconds on average, including CPU as-of lookup and answer assembly; it performs neither neural decoding nor visual replay. This read uses the already maintained target state.

A late-target diagnostic retains exactly \RuntimeReplayFrames{} observations from a \DurationLongMin{}-minute history. Replay uses these retained frames without adding an observation at the zero-time boundary. On target revelation, those cached frames are processed chronologically through the counting stream, using observations no later than the query cutoff. Replay processing time is the retained-frame count divided by \StreamIngestFPS{} frames per second; adding the typed-read cost gives \LateTargetReplaySeconds{} seconds after rounding. This cost calculation uses the explicit retained trace and differs from the Full general-QA request above. The host ledger occupies \LedgerShortKB{} and \LedgerLongKB{} kilobytes after \DurationShortMin{} and \DurationLongMin{} minutes, respectively, using decimal kilobytes.

%% file: sections/98_diagnostics.tex
\section{Precursor Component Evaluation}
\label{app:diagnostics}
The precursor continuous-memory experiments use Qwen3-VL-8B with parallel MLP-TTT branches and the language-model answer decoder. Full replays each causal prefix independently. Persistent Stream uses a local KV window, carries visual state between questions, and discards each QA fork after decoding. These experiments use nonlinear MLP fast weights and precede the delta-memory transition model. They identify effects of that continuous component.

\paragraph{Continuous-memory adaptation.}
The continuous branch admits a separate intervention before an event reader is attached. On the adaptation split, two-stage training increases Full GPA from \HistFullStageOne{} to \HistFullStageTwo{}. Setting the learned branch gates to zero reduces it to \HistFullGateZero{}; the paired difference is \HistFullBranchGain{} points, with a \bilingual{95\%}{95\%} interval of [\HistFullCILow{}, \HistFullCIHigh{}]. This establishes that the adapted model uses the branch, including its interaction with the jointly adapted language model.

On the development split, the dynamically updated streaming model scores \DevMFive{} GPA, compared with \DevStatic{} for a fixed fast-weight state. Initializing from the Full-adapted checkpoint raises the score to \DevFA{}, a paired gain of \DevFAgain{} points [\DevFACILow{}, \DevFACIHigh{}]. Further streaming training reaches \DevTZero{}; its additional GPA interval [\DevTCILow{}, \DevTCIHigh{}] crosses zero. These results motivate supervising state transitions directly rather than relying on additional recurrent training alone. The component settings and reader control are given below.

\begin{table}[h]
\centering\small
\caption{Continuous-component diagnostics. Full rows evaluate the adaptation split; Stream rows evaluate the development split. Each block retains its precision and training setup and is ranked separately.}
\label{tab:continuous}
\begin{tabular}{llc}
\toprule Access / population & Configuration & GPA $\uparrow$\\\midrule
Full / \HistTrainQueries{} queries & First-stage adaptation & \Second{\HistFullStageOne{}}\\
Full / \HistTrainQueries{} queries & Two-stage adaptation & \Best{\HistFullStageTwo{}}\\
Full / \HistTrainQueries{} queries & Two-stage, branch gates zero & \HistFullGateZero{}\\\midrule
Stream / \DevQueries{} queries & Recurrent initialization & \DevMFive{}\\
Stream / \DevQueries{} queries & Fixed fast weights & \DevStatic{}\\
Stream / \DevQueries{} queries & Full-adapted initialization & \Second{\DevFA{}}\\
Stream / \DevQueries{} queries & Further streaming adaptation & \Best{\DevTZero{}}\\\midrule
Stream / \DevQueries{} queries & Matched-precision reader control & \Second{\DevReaderBase{}}\\
Stream / \DevQueries{} queries & Reader adaptation & \Best{\DevReader{}}\\\bottomrule
\end{tabular}
\end{table}
Zeroing the trained branch gates measures dependence on the branch in the jointly adapted network. Freezing fast weights stops the update, momentum, and decay together. These interventions characterize branch use and recurrent dynamics; the discrete-state interface is evaluated separately.

The reader experiment freezes the continuous-memory parameters and fits a last-layer MLP adapter. Exact answers increase from \DevReaderBaseCorrect{} to \DevReaderCorrect{} out of \DevQueries{}, an EM gain of \ReaderEMGain{} points [\ReaderEMCILow{}, \ReaderEMCIHigh{}]. The GPA interval crosses zero. This distinction is expected because query-point EM and question-averaged GPA assign different weights to trajectories.

General video understanding is checked on \MMEQueries{} Video-MME questions without subtitles. Full-adapted initialization scores \MMEBase{}\%, while further streaming count adaptation scores \MMEAdapted{}\%. The task-specific adaptation therefore changes the accuracy tradeoff beyond counting.

\paragraph{Paired state-summary readout.}
A later paired control evaluates an external state-summary reader on the same \DevQueries{} queries. The memory condition answers \PairedMemoryCorrect{} correctly (\PairedMemoryEM{}\%), while zeroing its memory input yields \PairedZeroCorrect{} (\PairedZeroEM{}\%): a difference of \PairedNetEM{} points. Across \SaturationRecords{} training-cache records, \SaturationFraction{}\% of the inspected write activations are saturated and the minimum cosine similarity between read directions is \SaturationCosine{}. This adapter behaves almost like a fixed readout direction; its result does not establish content-specific long-term retrieval. It is distinct from the fast-weight gate test and the typed event-state interface.

\paragraph{Precursor execution.}
In a precursor MLP-memory execution test, the 8B engine processes \RuntimeFrames{} frames and \RuntimeTokens{} video tokens across \RuntimeQueries{} queries. It sustains \RuntimeAppend{} fps with \RuntimeGPU{} GiB GPU allocation, \RuntimeCPUlow{}--\RuntimeCPUhigh{} GiB host RSS, and \RuntimeLatency{} seconds p95 ready-prefix first-token latency.

\section{Count-State Readout}
\label{app:state-results}
\begin{table}[h]
\centering\small
\caption{Structured updates and count-state readout on SVCBench (GPA). The threshold variant is retrained; the remaining controls modify only the count projection read from a fixed complete-model rollout.}
\label{tab:state}
\begin{tabular}{lcc}
\toprule Condition & Full $\uparrow$ & Stream $\uparrow$\\\midrule
\rowcolor{OurRowGray}Complete model & \Best{\MatchedEightFull{}} & \Best{\MatchedEightStream{}}\\
Threshold updates in place of FSM & \Second{\NoFSMFull{}} & \Second{\NoFSMStream{}}\\
Clear count projection & \ZeroFull{} & \ZeroStream{}\\
Cross-video count projection & \SwapFull{} & \SwapStream{}\\
Previous-endpoint projection & \DelayFull{} & \DelayStream{}\\\bottomrule
\end{tabular}
\end{table}
The fixed-sensor assay uses a causal pixel recurrence estimator on \SensorParents{} periodic videos. Prefix estimates are computed at each query time, and the append-only ledger commits supported nonnegative increments. The direct estimates and the ledger both achieve \SensorExact{}\% exactness; the ledger prevents downward revisions and changes MAE from \SensorDirectMAE{} to \SensorLedgerMAE{}. Clearing and delaying the ledger each yield \SensorZeroExact{}\% exactness; cross-video swapping yields \SensorSwapExact{}\%. This assay holds perception fixed and measures the dependence of the readout on correctly indexed state.

%% file: sections/96_transfer.tex
\section{Online and Spatial Transfer Protocols}
\label{app:transfer}
\paragraph{Shared checkpoints and causal input.}
The transfer tests use the same counting-trained 4B and 8B checkpoints as SVCBench. No OVO-Bench or OVO-S questions, answers, or evaluation videos are used for task adaptation or checkpoint selection. A single training selection jointly excludes both benchmarks' matching source videos and transformed descendants. Full samples at most \FullFrameBudget{} prefix frames on OVO-Bench and \OVOSFullFrames{} on OVO-S. Stream ingests at \StreamFPS{} fps with a \StreamKVBudget{}-token window. The pretrained Stream controls retain the same observation rate and KV budget without the learned fast-weight or counting-state modules. Runtime state resets by video, while question decoding uses disposable forks. Benchmark prompts determine the original answer format; these general questions do not supply the typed counting interface.

\paragraph{OVO-Bench.}
The fixed released annotation set contains \OVOReleaseItems{} records from \OVOReleaseVideos{} videos, yielding \OVOReleaseQueries{} scored decisions. We retain real-time visual perception (RT), backward tracing (BT), and forward active responding (FA) \citep{niu2025ovo}. Accuracy is computed over queries within each task, then averaged over tasks within a group and over the three groups. FA retains every scheduled trigger in the accuracy track, including repeated timestamps belonging to separate entries. A Full decision replays its available prefix; a Stream decision advances through newly arrived frames. Trigger schedules and answer extraction are shared across models; invalid responses count as incorrect.

\paragraph{OVO-S-Bench.}
The released OVO-S snapshot contains \OVOSReleaseItems{} items from \OVOSReleaseVideos{} videos; expanding multi-time questions gives \OVOSReleaseQueries{} scored queries. Its four levels cover instantaneous egocentric perception, spatiotemporal context tracking, generative spatial reasoning, and allocentric mapping \citep{li2026ovosbench}. We retain the multiple-choice prompts, greedy decoding, and \OVOSOutputTokens{}-token output cap. Answer extraction accepts the labels in each question's option set; invalid or failed responses remain in the denominator. We compute query accuracy within each task family, average the three families in each level, and give the four levels equal weight in L-Avg. Query-pooled accuracy is retained separately. The same frozen query list and aggregation apply to every initial-model and adapted-model comparison.

\paragraph{Reference protocols.}
Scores for StreamForest-7B and the memory-based Flash-VStream-7B \citep{flashvstream2024} on OVO-Bench follow the three-group results in \citet{zeng2025streamforest}, with native sequential ingestion. Their spatial scores follow \citet{li2026ovosbench}. Both are grouped under Stream, with their original evaluation releases and input rates: StreamForest uses one frame per second on both benchmarks; Flash-VStream uses one frame per second on OVO-Bench and two on OVO-S. The matched Qwen controls and StaMina share the frozen evaluation releases, observation budgets, and answer extraction described above.

On the original \OVOSQuestions{}-item OVO-S evaluation set, Qwen3-VL-4B uses the same prefix-frame cap as Full and scores \OSSBaseFourFullLevelOne{}/\OSSBaseFourFullLevelTwo{}/\OSSBaseFourFullLevelThree{}/\OSSBaseFourFullLevelFour{} on L1--L4, with L-Avg.\ \OSSBaseFourFullAvg{} \citep{li2026ovosbench}. The Qwen3-VL-4B comparison in Table~\ref{tab:transfer-details} instead uses the frozen evaluation release specified above.

\TransferDetails
The two transfer benchmarks probe different question families and are not combined into a single score. Their source domains can overlap. The breakdown distinguishes improvements in historical and spatial context from changes in immediate perception and general reasoning; it does not assume uniform transfer across tasks. The Video-MME control in Appendix~\ref{app:diagnostics} additionally records the behavior of an earlier component-only adaptation recipe.

%% file: sections/97_additional.tex
\section{Additional Ablations and Model Scaling}
\label{app:additional}
\begin{figure}[tb]
\centering
\includegraphics[width=\linewidth]{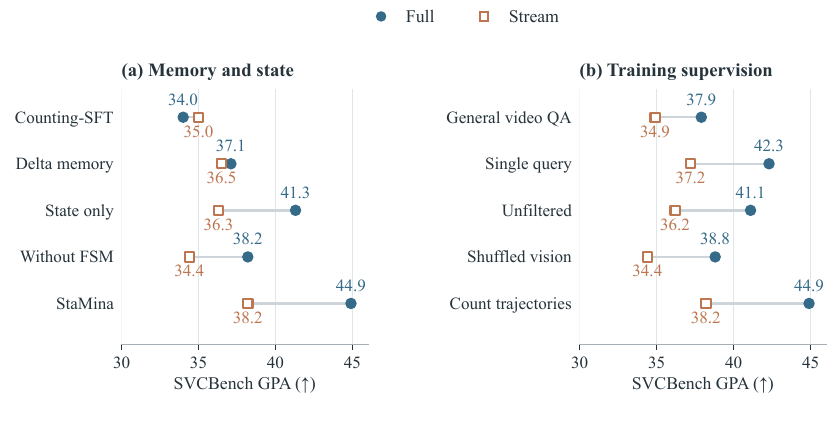}
\caption{System and supervision ablations at 8B on SVCBench. State only removes delta memory; without FSM replaces the finite-state machine with independent thresholded increments.}
\label{fig:ablations}
\end{figure}

\begin{figure}[!htbp]
\centering
\includegraphics[width=.80\linewidth]{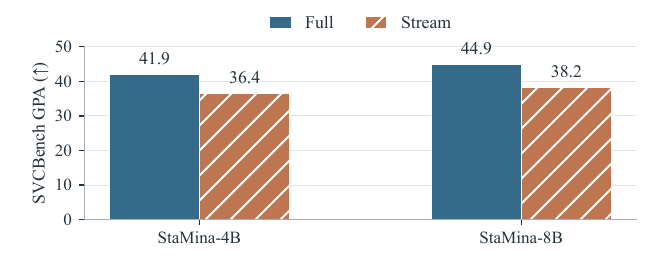}
\caption{Model scale and access on SVCBench. Each bar is the corresponding official-question aggregate in Table~\ref{tab:main}.}
\label{fig:scale}
\end{figure}

\begin{figure}[!htbp]
\centering
\includegraphics[width=.78\linewidth]{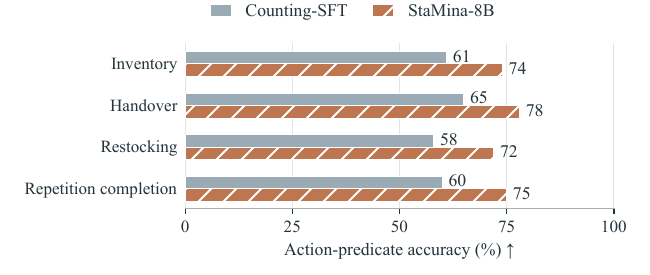}
\caption{Action-predicate accuracy on recorded task videos. Both 8B conditions use Stream with identical predicates and decision times.}
\label{fig:embodied}
\end{figure}

Figure~\ref{fig:ablations} summarizes system composition and supervision controls. These controls change modules or supervision, whereas Table~\ref{tab:mechanism} retains the legal graph to isolate transition learning.
\begin{table}[!htbp]
\centering\small
\caption{SVCBench system variants at 8B. Adapted rows share counting queries and update budgets; auxiliary objectives follow their components. Full and Stream share a checkpoint within each row. The two single-component variants are independent controls.}
\label{tab:paired}
\begin{tabular}{lcc}
\toprule Model & Full $\uparrow$ & Stream $\uparrow$\\\midrule
Qwen3-VL-8B & \MatchedEightOriginalFull{} & \MatchedEightOriginalStream{}\\
Counting-SFT & \MatchedEightSFTFull{} & \MatchedEightSFTStream{}\\
Counting-SFT + delta memory & \MatchedEightTTTFull{} & \Second{\MatchedEightTTTStream{}}\\
Counting-SFT + explicit state & \Second{\MatchedEightLedgerFull{}} & \MatchedEightLedgerStream{}\\
\rowcolor{OurRowGray}\StaMinaEight{} (Ours) & \Best{\MatchedEightFull{}} & \Best{\MatchedEightStream{}}\\\bottomrule
\end{tabular}
\end{table}
The data controls use the same 8B checkpoint family, training updates, and visual-token budget. The visual-alignment control keeps each trajectory's timestamp slots, legal count endpoints, and supervision masks fixed while permuting its visual observations. The candidate association graph is held fixed in this control, so count reachability and the number of supervised endpoints are unchanged. Observations are processed causally in the resulting perturbed sequence; this is a training corruption, not an alternative evaluation input. Filtering changes label selection. Figure~\ref{fig:scale} summarizes the access gap at both model scales, and Table~\ref{tab:embodied} gives the exact predicate accuracies underlying Figure~\ref{fig:embodied}.
\begin{table}[!htbp]
\begin{minipage}[t]{.48\linewidth}
\centering\footnotesize\setlength{\tabcolsep}{3pt}
\caption{Supervision ablations at matched token and update budgets.}
\label{tab:data}
\begin{tabular}{lcc}
\toprule Supervision & Full $\uparrow$ & Stream $\uparrow$\\\midrule
General video QA & \DataGenericFull{} & \DataGenericStream{}\\
Single query per trajectory & \Second{\DataSingleFull{}} & \Second{\DataSingleStream{}}\\
Before consistency filtering & \DataUnfilteredFull{} & \DataUnfilteredStream{}\\
Shuffled visual alignment & \DataShuffledFull{} & \DataShuffledStream{}\\
\rowcolor{OurRowGray}Complete count trajectories & \Best{\MatchedEightFull{}} & \Best{\MatchedEightStream{}}\\\bottomrule
\end{tabular}
\end{minipage}\hfill
\begin{minipage}[t]{.48\linewidth}
\centering\footnotesize\setlength{\tabcolsep}{3pt}
\caption{Action-predicate accuracy on recorded videos (\%, $\uparrow$). Each cell pools \PredicateTrials{} paired decisions; ranking is within each routine.}
\label{tab:embodied}
\begin{tabular}{lc>{\columncolor{OurRowGray}}c}
\toprule Routine & SFT & \shortstack{\StaMinaEight{}\\(Ours)}\\\midrule
Inventory & \Second{\PredicateInventoryBase{}} & \Best{\PredicateInventorySta{}}\\
Handover & \Second{\PredicateHandoverBase{}} & \Best{\PredicateHandoverSta{}}\\
Restocking & \Second{\PredicateRestockBase{}} & \Best{\PredicateRestockSta{}}\\
Repetition completion & \Second{\PredicateCycleBase{}} & \Best{\PredicateCycleSta{}}\\\bottomrule
\end{tabular}
\end{minipage}
\end{table}

